\documentclass{article}

\PassOptionsToPackage{authoryear, sort&compress}{natbib}

\usepackage[final]{neurips_2024}

\usepackage[colorinlistoftodos,prependcaption,textsize=tiny]{todonotes}
\usepackage{xargs} 
\usepackage{wrapfig}
\usepackage{threeparttable}
\usepackage{enumitem}
\usepackage{makecell}
\usepackage[utf8]{inputenc} %
\usepackage[T1]{fontenc}    %
\usepackage{url}            %
\usepackage{booktabs}       %
\usepackage{amsfonts}       %
\usepackage{nicefrac}       %
\usepackage{xcolor}         %
\usepackage{colortbl}       %
\usepackage{microtype}
\usepackage{graphicx}
\usepackage{floatrow}
\usepackage{subcaption}
\usepackage{xurl} 
\usepackage{listings}
\usepackage{fancyvrb}
\usepackage{longtable}
\usepackage{array}
\usepackage[breakable]{tcolorbox} %
\usepackage{multirow}
\usepackage{marvosym}
\usepackage{fontawesome5}  %

\definecolor{titleblue}{RGB}{25, 70, 135}
\definecolor{linkblue}{RGB}{50, 120, 200}
\definecolor{iconred}{RGB}{200, 60, 60}
\definecolor{iconpurple}{RGB}{120, 80, 160}
\definecolor{iconblue}{RGB}{60, 100, 180}
\definecolor{icongold}{RGB}{205, 130, 15}
\definecolor{iconteal}{RGB}{20, 145, 135}
\definecolor{boxblue}{RGB}{230, 242, 255}  %

\newtcolorbox{titlebox}{
    colback=boxblue,
    colframe=boxblue,
    arc=4pt,
    boxrule=0pt,
    left=12pt,
    right=12pt,
    top=10pt,
    bottom=10pt,
    width=\textwidth,
}

\newcommand{\customtitle}[1]{%
    \begin{center}
    {\fontsize{18}{22}\selectfont\bfseries\color{titleblue}#1}
    \end{center}
    \vspace{0.3cm}
}

\newcommand{\corresp}{\Letter}        %

\newcommand{\customauthors}[1]{%
    \begin{center}
    #1
    \end{center}
    \vspace{0.1cm}
}

\newcommand{\customaffiliations}[1]{%
    \begin{center}
    #1
    \end{center}
}

\newenvironment{customabstract}{%
    \vspace{0.2cm}
    \noindent\textbf{Abstract:}~%
}{%
    \par\vspace{0.3cm}
}

\newcommand{\keywords}[1]{%
    \noindent\textbf{Keywords:}~~#1
    \vspace{0.3cm}
}

\newcommand{\paperdate}[1]{%
    \noindent{\fontsize{8.5}{10}\selectfont\textcolor{iconred}{\makebox[1.15em][c]{\faIcon{calendar-alt}}}~\textbf{Date:}~#1}
}

\newcommand{\githubrepo}[1]{%
    \noindent{\fontsize{8.5}{10}\selectfont\textcolor{iconpurple}{\makebox[0.88em][c]{\faIcon{github}}}~\textbf{Code:}~\href{#1}{\textcolor{linkblue}{#1}}}
}

\newcommand{\hfmodel}[1]{%
    \noindent{\fontsize{8.5}{10}\selectfont\textcolor{icongold}{\makebox[0.88em][c]{\faIcon{cube}}}~\textbf{Model:}~\href{#1}{\textcolor{linkblue}{#1}}}
}

\newcommand{\hfdata}[1]{%
    \noindent{\fontsize{8.5}{10}\selectfont\textcolor{iconteal}{\makebox[0.88em][c]{\faIcon{database}}}~\textbf{Data:}~\href{#1}{\textcolor{linkblue}{#1}}}
}

\newcommand{\contactemail}[1]{%
    \noindent{\fontsize{8.5}{10}\selectfont\textcolor{iconblue}{\makebox[1.15em][c]{\faIcon{envelope}}}~\textbf{Contact:}~#1}
}

\usepackage[bookmarks=true,colorlinks=true,linkcolor=blue!60!black,urlcolor=magenta,citecolor=green!40!black]{hyperref}

\usepackage{amsmath}
\usepackage{amssymb}
\usepackage{mathtools}
\usepackage{amsthm}
\theoremstyle{plain}

\theoremstyle{definition}

\theoremstyle{remark}

\usepackage{bm} 
\usepackage{algorithm, algorithmic}

\usepackage{etoolbox}
\usepackage{natbib}
\newcounter{bibcount}
\makeatletter
\patchcmd{\@lbibitem}{\item[}{\item[\hfil\hspace{2.5em}\stepcounter{bibcount}{[\thebibcount]}\hspace{0.em}}{}{}
\makeatother
\usepackage[nameinlink,capitalise]{cleveref}

\usepackage{titletoc}
\crefname{section}{§}{§§}
\Crefname{section}{§}{§§}

\usepackage{pifont}

\definecolor{softblue}{RGB}{100, 130, 230}    %
\definecolor{softcoral}{RGB}{200, 100, 110}   %
\definecolor{softred}{RGB}{200, 90, 90}       %
\definecolor{softgreen}{RGB}{90, 160, 110}    %

\definecolor{impgreen}{RGB}{60, 140, 90}  %

\definecolor{bestcolor}{RGB}{183, 223, 235}     %
\definecolor{secondcolor}{RGB}{225, 242, 248}   %
\definecolor{offlinecolor}{RGB}{232, 245, 233}  %
\definecolor{onlinecolor}{RGB}{255, 243, 224}   %

\definecolor{promptframe}{RGB}{185, 204, 226}
\definecolor{prompttitlebg}{RGB}{210, 226, 245}
\definecolor{prompttitlefg}{RGB}{25, 55, 100}
\newtcolorbox{promptbox}[1][]{
    breakable,
    colback=boxblue,
    colframe=promptframe,
    arc=2.5pt,
    boxrule=0.5pt,
    left=8pt,
    right=8pt,
    top=6pt,
    bottom=6pt,
    fontupper=\small\ttfamily,
    colbacktitle=prompttitlebg,
    coltitle=prompttitlefg,
    fonttitle=\small\bfseries,
    toptitle=2.5pt,
    bottomtitle=2.5pt,
    #1
}

\usepackage{placeins}
\newcommand{\chironfirstpagebegin}{%
  \savegeometry{chironbody}%
  \newgeometry{left=14mm,right=14mm,top=14mm,bottom=14mm,includefoot,footskip=0mm,headheight=0pt,headsep=0pt}%
  \begingroup
  \setlength{\intextsep}{2pt}%
  \setlength{\textfloatsep}{2pt}%
  \setlength{\floatsep}{2pt}%
}
\newcommand{\chironfirstpageend}{%
  \par\FloatBarrier
  \endgroup
  \loadgeometry{chironbody}%
}
\newcommand{\chironcompactarxiv}{%
  \renewtcolorbox{titlebox}{colback=boxblue,colframe=boxblue,arc=3pt,boxrule=0pt,left=\dimexpr12pt-2mm\relax,right=\dimexpr12pt-2mm\relax,top=12pt,bottom=12pt,width=\textwidth,leftright skip=2mm,before skip=0pt,after skip=2pt}%
  \renewcommand{\customtitle}[1]{\par{\centering\huge\bfseries\color{titleblue}##1\par}\vspace{8pt}}%
  \renewenvironment{customabstract}{\par\fontsize{11}{13}\selectfont\noindent\textbf{Abstract:}~}{\par\vspace{6pt}}%
  \renewcommand{\keywords}[1]{\par\noindent{\fontsize{8.5}{10}\selectfont\textbf{Keywords:}~##1\par}\vspace{8pt}}%
  \renewcommand{\paperdate}[1]{\noindent{\fontsize{8.5}{10}\selectfont\textcolor{iconred}{\faIcon{calendar-alt}}~\textbf{Date:}~##1}}%
  \renewcommand{\contactemail}[1]{\noindent{\fontsize{8.5}{10}\selectfont\textcolor{iconblue}{\faIcon{envelope}}~\textbf{Contact:}~##1}}%
}

\newcommand{\chiron}{Chiron}
\newcommand{\R}{\mathbb{R}}

\begin{document}
\chironfirstpagebegin
\chironcompactarxiv

\thispagestyle{empty}

\begin{titlebox}
\customtitle{A Foundation Model\\[6pt]to Measure and Predict Human Behaviour}

\customauthors{%
\textbf{Haoran Ye}\textsuperscript{1,*}\quad
\textbf{Tianrui Huo}\textsuperscript{2,*}\quad
\textbf{Hangzhen Li}\textsuperscript{2}\quad
\textbf{Xin Zhang}\textsuperscript{3}\quad
\textbf{Guojie Song}\textsuperscript{1,\,\corresp}\\[0.2cm]
{\textsuperscript{*}Equal contribution}
}

\customaffiliations{%
\textsuperscript{1}State Key Laboratory of General Artificial Intelligence, School of Intelligence Science and Technology, Peking University\\
\textsuperscript{2}Yuanpei College, Peking University\\
\textsuperscript{3}School of Psychological and Cognitive Sciences, Peking University
}
\vspace{0.2cm}

\begin{customabstract}
Large language models (LLMs) are increasingly used to infer psychological attributes, predict individual behaviour, and simulate human populations. However, despite generating plausible responses, they often misrepresent individuals and distort population statistics. Here we introduce \chiron{}, a foundation model to measure and predict human behaviour. We jointly fine-tune an LLM to represent psychological constructs, predict individual behaviour, and align population response distributions with human statistics. The training and evaluation data span 36 behavioural domains and 78 social science experiments, comprising 288,307 person records, 1,411,750 atomic behavioural decisions, and 99,953 psychometric assessments over 109 psychological constructs.
In our evaluations, \chiron{}'s psychological measurements agree more closely with human self-reports and exhibit fewer biases that undermine psychometric validity.
It predicts individual behaviour more accurately than frontier LLMs (up to 28.4\% relative improvement). 
In simulations of social science experiments, it more accurately captures the direction and magnitude of human experimental effects 
(82.2--94.3\% lower effect-size estimation error).
Together, these results establish \chiron{} as a unified computational instrument for measuring and predicting human behaviour.

\end{customabstract}

\noindent
\begin{minipage}[b]{0.78\textwidth}
\githubrepo{https://github.com/ValueByte-AI/chiron}\\[0.05cm]
\hfmodel{https://huggingface.co/chiron-group/chiron}\\[0.05cm]
\hfdata{https://huggingface.co/datasets/chiron-group/chiron-data}\\[0.14cm]
\paperdate{November, 2026}\quad\contactemail{\href{mailto:hrye@stu.pku.edu.cn}{\textcolor{linkblue}{hrye@stu.pku.edu.cn}} \quad \href{mailto:gjsong@pku.edu.cn}{\textcolor{linkblue}{gjsong@pku.edu.cn}}}
\end{minipage}%
\hfill
\begin{minipage}[b]{0.18\textwidth}
\raggedleft
\includegraphics[height=1.5cm]{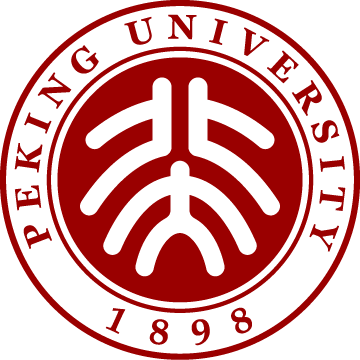}
\end{minipage}
\end{titlebox}

\begin{figure}[!b]
\centering
\makebox[\linewidth][c]{%
\begin{minipage}{182mm}
\centering
\includegraphics[width=\linewidth,trim=0 17mm 0 0,clip]{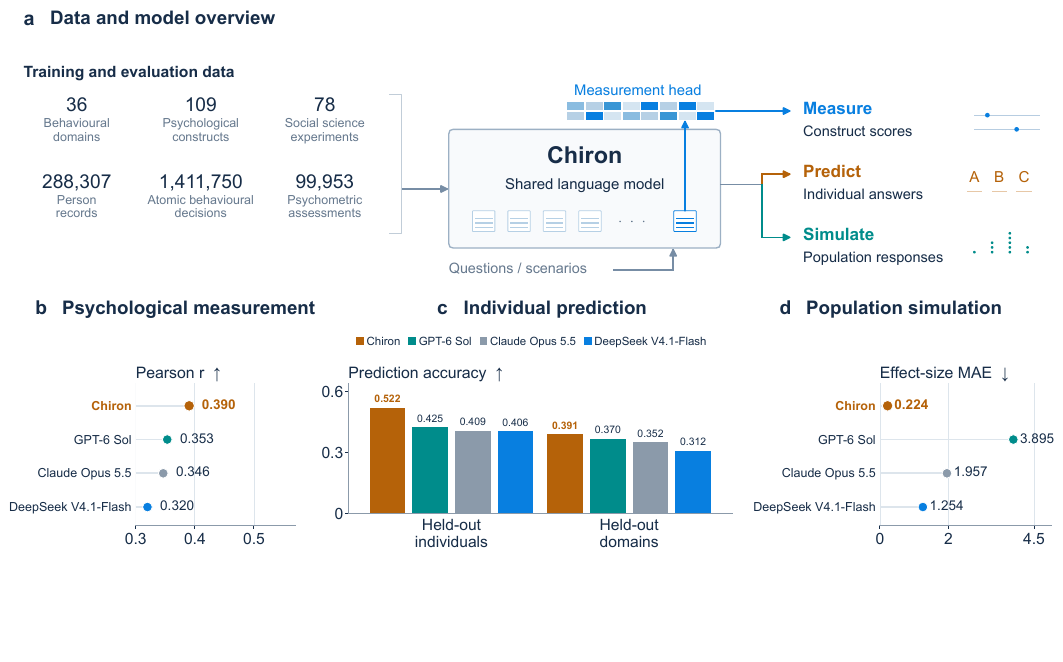}\par
\caption{\textbf{Chiron: data, model, and main comparisons.} \textbf{a}, Training and evaluation coverage and the shared model, which measures psychological constructs, predicts individual behaviour, and simulates populations. \textbf{b}, Psychological measurement: mean Pearson correlation against self-report assessments across five datasets (nine constructs). \textbf{c}, Individual behavioural prediction: mean prediction accuracy across 29 behavioural domains. \textbf{d}, Population simulation: Cohen's $d$ MAE across 25 eligible effects from eight studies.}
\label{fig:headline-results}
\end{minipage}%
}
\end{figure}
\FloatBarrier
\chironfirstpageend

\section{Introduction}\label{sec:intro}

A central goal of psychology and the behavioural sciences is to understand how people differ and how these differences shape behaviour across situations \citep{cronbach1955construct}. Psychological measurement draws on what people say and do to infer latent attributes such as personality traits, values, and affective states, making subjective inner experience explicit and comparable across individuals \citep{demszky2023using}. Behavioural prediction uses the available evidence to anticipate how individuals and populations will respond in new situations. Together, measurement and prediction provide complementary foundations for modelling human behaviour, linking interpretable psychological constructs with empirical predictive power \citep{hofman2021integrating}.

Recent work has begun to use large language models (LLMs) across these tasks, deploying them both as analytical instruments to infer psychological attributes from digital traces \citep{peters2024dispositions,wright2026assessment,ziems2024can} and as simulated research subjects to predict behaviour and replicate surveys or experiments \citep{argyle2023silicon,aher2023turing}.
However, off-the-shelf LLMs often generate fluent, plausible responses that misrepresent individual differences and distort population statistics, raising several foundational challenges across psychology and computational social science.
(1) \emph{Unstructured psychological assessments suffer from systematic psychometric biases.} When prompted to score psychological attributes from text, general-purpose LLMs inflate socially desirable traits, compress inter-individual score dispersion, and shift estimates under superficial changes in prompt framing \citep{abdurahman2024perils,ye2025llmpsychometrics,brickman2025assessment}.
(2) \emph{Psychological construct interpretability is disconnected from individual behavioural prediction.} Existing approaches either extract trait scores without verifying out-of-sample predictive power \citep{zhu2025personality} or fine-tune black-box predictors without grounding representations in interpretable constructs \citep{binz2025centaur,xie2025befm}. Moreover, unadapted LLMs struggle to infer individuals' distinctive values from their statements \citep{jiang2025indievalue}, flatten within-group diversity into demographic stereotypes \citep{wang2024misportray}, and compress the linguistic diversity that reflects different individual traits \citep{sourati2026diversity}.
(3) \emph{Simulated populations exhibit homogenization and variance collapse.} At the population level, LLM-based simulations cluster around stereotypical or modal responses \citep{kozlowski2025simulating}. While simulated cohorts often recover the qualitative direction of experimental treatments, they systematically underestimate within-condition human variance. Consequently, these simulations severely exaggerate standardized effect magnitudes \citep{ashokkumar2026experiments}, often yield false positives on null findings \citep{cui2025largescale}, and only weakly capture cross-national variations in human preferences \citep{chen2026preferences}.
(4) \emph{Privacy and data-access constraints limit individual-level calibration.} Adapting models to overcome these distortions requires rich behavioural supervision. However, raw human records paired with comprehensive psychometric annotations or participant-level experimental responses are rarely accessible at scale due to privacy and institutional constraints \citep{xu2026llmcss}.

These failures stem from a fundamental mismatch between the training objectives of general-purpose LLMs and the methodological requirements of behavioural science. Standard instruction tuning and preference optimization reward responses judged helpful, normative, or socially desirable by annotators, but do not require models to anchor internal representations in psychometric constructs, preserve idiosyncratic individual differences, or calibrate output distributions to human population variance. Aggregating feedback across annotators further privileges modal preferences at the expense of minority and individual perspectives \citep{santurkar2023whose,sorensen2024roadmap,conitzer2024social}. Consistent with this mechanism, reinforcement learning from human feedback (RLHF) reduces output entropy \citep{kirk2024rlhf} and collapses conceptual diversity in simulated populations \citep{murthy2025onefish}. Addressing the challenges above therefore requires moving beyond zero-shot prompting of conversational chatbots toward a foundation model explicitly architected and trained for psychometric validity, individual predictive fidelity, and distributional alignment.

\begin{figure}[t]
\centering
\includegraphics[width=\linewidth]{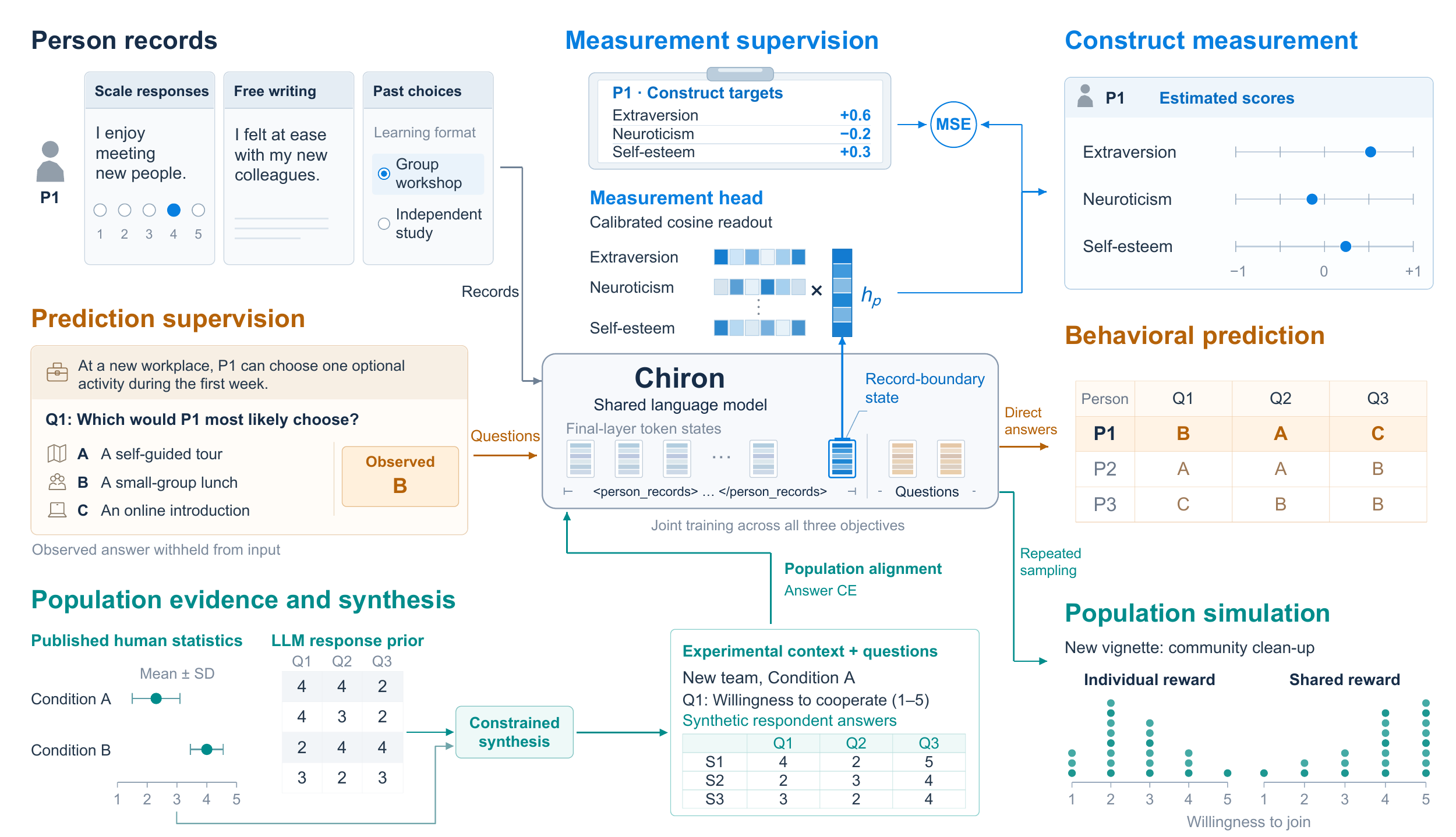}
\caption{Overview of \chiron{}. A single language model is jointly trained across three objectives: (1) \emph{psychological measurement}, where construct scores supervise a measurement head to estimate psychological attributes from personal records; (2) \emph{individual behavioural prediction}, where answers withheld from a person's records supervise the prediction of unseen responses and choices; and (3) \emph{population simulation}, where virtual respondents synthesized to match published human statistics supervise the generation of population-level response distributions.}
\label{fig:overview}
\end{figure}

Here we introduce \chiron{}, a foundation model to measure and predict human behaviour (\cref{fig:headline-results,fig:overview}). \chiron{} jointly learns psychological measurement, individual behavioural prediction, and population alignment within a shared language model through three mechanisms.
(1) \emph{A decoupled latent readout estimates psychological constructs directly from person representations.} \chiron{} attaches a centred cosine measurement head to the hidden state at the end of a person's observed records, scoring 109 psychological constructs without verbalized text generation and avoiding generative biases.
(2) \emph{Joint psychometric supervision anchors individual behavioural prediction in latent construct representations.} Measurement supervision across the 109 constructs directly shapes the shared person representation. The model then predicts individual behaviour from this structured psychological representation, linking interpretable traits with accurate behavioural prediction.
(3) \emph{Constrained population synthesis calibrates the model's response distribution to human experimental statistics.} Because participant-level experimental data are rarely public, we synthesize virtual respondent cohorts whose answers match published condition means, standard deviations, and treatment effects. By training the model to generate diverse cohort responses for each experimental condition, \chiron{} counteracts variance collapse and reproduces human response distributions.

We train \chiron{} on a heterogeneous collection of questionnaire responses, personal writing, online posts, and experimental data. The combined training and evaluation data span 36 behavioural and psychological assessment domains and 78 social science experiments, comprising 288{,}307 person records, 1{,}411{,}750 atomic behavioural decisions and 99{,}953 psychological scores spanning 109 psychological constructs. To overcome privacy and data-access barriers where raw human records or comprehensive annotations are unavailable, our training framework combines direct supervision from human records and psychometric assessments, self-supervised masked observation prediction on unlabelled personal writing, and calibrated synthetic supervision derived from psychometric item pools and published aggregate human statistics.

We evaluate \chiron{} across psychological measurement, individual behavioural prediction, and population simulation. 
Its psychological measurements agree more closely with human self-reports across five datasets spanning nine constructs (mean Pearson $r=0.390$, compared with $0.320$--$0.353$ for frontier LLMs), while recovering empirical inter-construct correlations reported in the psychometric literature ($r=0.806$--$0.819$ across 150 construct pairs).
Compared with frontier LLMs, \chiron{} substantially mitigates measurement biases (including score compression, irrelevant-context sensitivity, and evidence-order sensitivity), and responds monotonically to experimental manipulations of construct-relevant evidence across all 109 constructs.
For individual prediction, \chiron{} predicts individual behaviour more accurately than frontier LLMs across 29 behavioural domains (mean prediction accuracy of $0.504$ versus $0.393$--$0.418$, up to 28.4\% relative improvement), generalizing both to unseen individuals and to held-out domains excluded from prediction training.
For population simulations across 15 held-out experimental studies, \chiron{} preserves realistic within-condition dispersion and reproduces human experimental effects with 90.5\% direction agreement and 82.2--94.3\% lower effect-size estimation error (Cohen's $d$ MAE of $0.224$ versus $1.254$--$3.895$ for frontier LLMs).
Together, these findings establish \chiron{} as a unified computational instrument for measuring psychological attributes and predicting human behaviour at individual and population scales.

\section{Related work}\label{sec:related}

Large language models are reshaping psychology and computational social science across multiple research roles, most notably as analytical assistants that extract psychological constructs from unstructured digital traces and as simulated research subjects that model individual decisions and collective experimental outcomes \citep{demszky2023using,ziems2024can,xu2026llmcss}. Below, we situate \chiron{} within prior work on psychological measurement, individual behavioural prediction, and population simulation, highlighting how our unified training framework addresses recurring validity and fidelity challenges across these areas.

\paragraph{Psychological measurement.}
Psychological measurement relates observable behavioural or linguistic evidence to latent constructs such as personality traits, values, and affective states \citep{cronbach1955construct}. Moving beyond traditional self-report questionnaires and closed-vocabulary lexicon tools, recent studies prompt or adapt LLMs to infer dispositions from social-media posts, essays, and clinical narratives \citep{peters2024dispositions,wright2026assessment}.
However, psychometric evaluations reveal substantial validity limitations in off-the-shelf LLMs: inferred scores often show modest agreement with self-reports \citep{zhu2025personality}, exhibit systematic positivity and Western-centric biases \citep{abdurahman2024perils,xu2026llmcss}, compress inter-individual variance \citep{ye2025llmpsychometrics}, and shift under superficial changes in prompt wording, surrounding context, or evidence ordering \citep{brickman2025assessment,ye2025foundationpsychometrics}.
Whereas general LLMs rely on autoregressive text generation to verbalize numerical ratings, \chiron{} introduces a dedicated centred cosine readout head trained on multi-source psychometric supervision across 109 constructs, decoupling construct measurement from generative reporting biases.

\paragraph{Individual behavioural prediction.}
A growing literature seeks to predict how specific individuals will think, choose, or act across situations, which is a prerequisite for reconciling explanatory psychological constructs with out-of-sample predictability \citep{hofman2021integrating,xu2026llmcss}. One approach supplies a general-purpose LLM with personal context in the prompt: generative agents condition behaviour on memory streams \citep{park2023generative}, and interview-grounded agents use qualitative transcripts to emulate specific individuals \citep{park2024agents1000}. 
While rich personal context improves simulation plausibility, unadapted LLMs still struggle to infer an individual's underlying values and preferences from their own statements \citep{jiang2025indievalue} and often fall back on demographic or persona stereotypes \citep{wang2024misportray,xu2026llmcss}.
A complementary line of work fine-tunes foundation models directly on behavioural data: Centaur learns from trial-by-trial choices across cognitive psychology experiments \citep{binz2025centaur}, and Be.FM integrates experimental, survey, and literature data for behavioural prediction \citep{xie2025befm}. \chiron{} advances this paradigm in three ways: it generalizes prediction across heterogeneous personal records (combining questionnaires, essays, and social-media traces), introduces self-supervised \emph{masked observation prediction} to leverage unlabelled personal writing at scale, and jointly supervises psychological measurement and behavioural prediction within a shared representation space so that latent construct structure informs individual choice prediction.

\paragraph{Population simulation.}
LLMs are also widely used as in silico research subjects to simulate survey populations \citep{argyle2023silicon}, replicate human-subject experiments \citep{aher2023turing,cui2025largescale,ashokkumar2026experiments}, and probe strategic or social interactions \citep{akata2025games,kozlowski2025simulating,xu2026llmcss}. Despite their promise for pretesting experimental designs and exploring counterfactual scenarios, empirical audits consistently show that off-the-shelf LLMs suffer from variance compression and hyper-accuracy distortion \citep{kozlowski2025simulating,xu2026llmcss}: simulated samples suppress within-group heterogeneity \citep{santurkar2023whose,wang2024misportray}, inflate standardized treatment effect sizes, produce false positives on null findings \citep{cui2025largescale,ashokkumar2026experiments}, and poorly reflect cross-national preference distributions \citep{chen2026preferences}.
Turning LLM simulations into reliable scientific instruments requires empirical calibration against observed human distributions, yet individual-level human responses are frequently unavailable due to privacy and institutional constraint \citep{xu2026llmcss}. \chiron{} addresses this bottleneck through \emph{constrained population synthesis}, reconstructing respondent-level supervision from published aggregate human statistics so that the model's conditional output distribution is directly aligned with empirical human means, variances, and treatment effects.

\section{Methods}\label{sec:method}

To address the methodological challenges of using language models in behavioural science (e.g., construct invalidity, loss of individual heterogeneity, and population variance collapse), \chiron{} unifies psychological measurement, individual behavioural prediction, and population simulation within a single foundation model (\cref{fig:overview}). We first describe the multi-source data corpus and task formulations (\cref{sec:method-data}), then detail the model architecture (\cref{sec:method-arch}) and joint training framework (\cref{sec:method-framework}), and finally discuss the mechanistic and methodological advantages of \chiron{} over both general-purpose LLMs and traditional psychological instruments (\cref{sec:method-advantages}).

\subsection{Data corpus and task formulation}\label{sec:method-data}

Human psychology and behaviour manifest across diverse empirical records: free-form personal writing, standardized psychometric inventories, everyday behavioural choices, and controlled experimental manipulations. We curate a unified training and evaluation corpus spanning 36 behavioural and psychological assessment domains (109 psychological constructs) and 78 scenario-based social science experiments, totalling 288{,}307 person records, 1{,}411{,}750 atomic behavioural decisions, and 99{,}953 psychometric assessments. As summarized with representative examples in \cref{tab:method-data-overview} (full dataset-by-dataset inventory in \cref{tab:training-data-detailed}), the corpus is organized around three complementary tasks.

\begin{table}[!t]
\centering
\caption{\textbf{Summary of training and evaluation data, task formulations, and illustrative examples across \chiron{}'s three training objectives.} Full dataset descriptions, split units, and all 109 construct definitions are provided in \cref{app:data} (\cref{tab:training-data-detailed,tab:data-population,tab:constructs}).}
\label{tab:method-data-overview}
\begingroup\fontsize{7.5}{9.2}\selectfont\setlength{\tabcolsep}{5pt}
\resizebox{\linewidth}{!}{%
\begin{tabular}{@{}
>{\raggedright\arraybackslash}p{\dimexpr0.20\linewidth-1.33\tabcolsep\relax}
>{\raggedright\arraybackslash}p{\dimexpr0.30\linewidth-1.33\tabcolsep\relax}
>{\raggedright\arraybackslash}p{\dimexpr0.50\linewidth-1.34\tabcolsep\relax}@{}}
\toprule
\textbf{Task \& data category} & \textbf{Representative sources \& constructs} & \textbf{Formulation (Input $\to$ Target) \& concrete example} \\
\midrule
\multicolumn{3}{@{}l@{}}{\textbf{1. Psychological measurement}} \\
\multicolumn{3}{@{}l@{}}{\emph{Scale:} 53{,}073 train records, 10{,}820 eval records; 99{,}953 psychometric assessments across 109 constructs} \\
\midrule
Human writing with self-reports & COVID-19 RW3D (anger, anxiety), Buechel empathy, PANDORA (Big Five), PsihoRo (PHQ-9 depression, GAD-7 anxiety), Student-SADD (PHQ-9 depression), myPersonality (Big Five, train only) & \textbf{Input $\mathcal{R}_p$:} Author's essays, reflections, or Reddit posts.\newline \textbf{Target $t_{pc}\in[-1,1]$:} Author's self-reported scale score.\newline \emph{Example:} ``I have barely slept all week worrying whether my parents will catch the virus...'' $\to$ Anxiety $t_{pc}=+0.75$. \\
\addlinespace[2pt]
Human questionnaires & 12 instruments: 16PF, AS+SC+AD+DO, Big Five, IPIP-FFM, HEXACO, DASS, ECR, RSE, SD3, MACH-IV, RIASEC, HSNS+DD & \textbf{Input $\mathcal{R}_p$:} Observed subset of a respondent's questionnaire answers.\newline \textbf{Target $t_{pc}\in[-1,1]$:} Construct score computed from withheld items on the questionnaire.\newline \emph{Example:} Observed responses on openness and agreeableness items $\to$ Extraversion score on withheld items ($t_{pc}=+0.40$). \\
\addlinespace[2pt]
Synthetic questionnaires \& calibrated narratives & All 109 constructs across 5 families (IPIP, HEXACO, DASS, ERQ, IRI, Schwartz values, MFQ-2 morality, Hofstede, BIS/BAS, NFC) & \textbf{Input $\mathcal{R}_p$:} Simulated item answers or first-person narratives controlled for direction, evidence strength, setting, and length.\newline \textbf{Target $t_{pc}\in[-1,1]$:} Instrument score or Bradley--Terry score.\newline \emph{Example:} Narrative describing reorganizing a team schedule after a setback $\to$ Conscientiousness $t_{pc}=+0.62$. \\
\addlinespace[2pt]
Nomological validity cohorts (evaluation only) & PAN15 social media, Pennebaker Essays, European Social Survey (ESS09); 150 literature construct pairs (60 constructs) & \textbf{Input $\mathcal{R}_p$:} Unlabelled posts, essays, or survey responses for each person in a cohort.\newline \textbf{Target:} Published correlation between construct pairs in the psychological literature (evaluated against correlations of \chiron{}-inferred scores across cohort members).\newline \emph{Example:} Inferred Depression and Anxiety correlate positively across individuals ($r=+0.65$), matching published benchmarks. \\
\midrule
\multicolumn{3}{@{}l@{}}{\textbf{2. Individual behavioural prediction}} \\
\multicolumn{3}{@{}l@{}}{\emph{Scale:} 192{,}179 train records (1{,}073{,}073 decisions) in 25 domains; 4{,}767 eval cases (30{,}773 decisions) in 29 domains} \\
\midrule
Psychometric \& attitude surveys & 19 training domains (e.g., 16PF, IPIP-FFM, DASS, ECR, CFCS, FTI, NPI, OSRI-44, World Values Survey) + 4 held-out domains (HEXACO, MACH-IV, RIASEC, SAPA-2022) & \textbf{Input $(\mathcal{R}_p,\mathbf{q})$:} Person's observed item responses + held-out questions.\newline \textbf{Target $\mathbf{y}$:} Person's actual withheld choices/ratings.\newline \emph{Example:} Observed WVS attitudes on family and work + ``Q: How justifiable is tax cheating? (A: Never ... J: Always)'' $\to$ \texttt{A}. \\
\addlinespace[2pt]
Personal writing with behavioural annotations & PANDORA Reddit, Essays Big Five, GRIT Tweets, LoST self-esteem, Twitter ST-MHD & \textbf{Input $(\mathcal{R}_p,\mathbf{q})$:} Author's posts/essays + condition or attribute classification query.\newline \textbf{Target $\mathbf{y}$:} Author's recorded mental-health condition, grit response, or categorical label.\newline \emph{Example:} Author's timeline of everyday tweets + ``Q: Which mental-health condition is associated with this user? (ADHD / OCD / PTSD / ...)'' $\to$ \texttt{ADHD}. \\
\addlinespace[2pt]
Naturalistic text (masked observation prediction) & Blog Authorship Corpus (unlabelled personal blogs) & \textbf{Input $(\mathcal{R}_p,\mathbf{q})$:} Extracted first-person behavioural observations + multiple-choice question on a withheld observation.\newline \textbf{Target $\mathbf{y}$:} Withheld behavioural tendency.\newline \emph{Example:} Visible habits + ``What activity do I prefer on weekends? (A: Solo reading ... C: Team sports)'' $\to$ \texttt{C}. \\
\midrule
\multicolumn{3}{@{}l@{}}{\textbf{3. Population alignment \& simulation}} \\
\multicolumn{3}{@{}l@{}}{\emph{Scale:} 23{,}191 train and 4{,}277 eval respondents (269{,}475 train + 38{,}429 eval = 307{,}904 decisions) across 78 experiments (\emph{AMJ}, \emph{JAP}, \emph{JEP}, \emph{JPSP}, \emph{OBHDP})} \\
\midrule
Social science experiments & 63 training studies (247 conditions) and 15 held-out test studies (45 conditions; 25 Cohen's $d$ effects, 42 direction calls, 53 significance tests, 34 distributions) & \textbf{Input $x$:} Experimental condition vignette and dependent-measure items.\newline \textbf{Target $\widetilde{\mathbf{y}}$:} Virtual respondent answers matching human condition means, SDs, and treatment effects ($d$).\newline \emph{Example:} Workplace negotiation framing (Condition A vs.\ B) $\to$ Sample $N_c$ independent completions to estimate group distribution and Cohen's $d$. \\
\bottomrule
\end{tabular}%
}
\endgroup
\end{table}

The 109 supervised psychological constructs (\cref{tab:constructs}) are selected from established psychometric frameworks to provide broad coverage across five domains of individual variation:
(1) \emph{broad personality and vocational interests} (24 constructs, including Big Five, HEXACO honesty--humility and emotionality, Short Dark Triad, and Holland's RIASEC vocational interests);
(2) \emph{self-evaluation, interpersonal attachment, and social functioning} (22 constructs, such as Rosenberg self-esteem, ECR attachment avoidance, and Interpersonal Reactivity Index empathy facets);
(3) \emph{affect, mental health, and wellbeing} (22 constructs, including DASS depression, anxiety, and stress, ERQ cognitive reappraisal and expressive suppression, resilience, and life satisfaction);
(4) \emph{values, morality, culture, and worldviews} (23 constructs, spanning Schwartz's basic human values, Moral Foundations Questionnaire-2 dimensions, and Hofstede value orientations); and
(5) \emph{cognition, self-regulation, and agency} (18 constructs, including need for cognition, BIS/BAS drive and reward responsiveness, impulse control, self-efficacy, and locus of control).
All individual-level evaluation sets are strictly disjoint from training data by participant identity where available and by source record otherwise (\cref{app:data-overview}).

\subsection{Model architecture}\label{sec:method-arch}

\chiron{} combines a shared causal language model backbone with a dedicated psychometric measurement head (\cref{fig:overview}). For individual-level tasks, the model takes a person's records as input, enclosed within \texttt{<person\_records>}...\texttt{</person\_records>}. The measurement head reads continuous construct scores directly from the resulting person representation, and the language model generates answers to subsequent prediction questions about that person. For population simulation, the language model takes an experimental scenario and its outcome questions as input and generates respondent-level answers via repeated sampling.

Let $\mathcal{R}_p=\{r_1,\ldots,r_n\}$ denote the records available for person $p$, which may consist of questionnaire item--response pairs, free-form personal writing, social-media posts, or recorded behavioural choices. We extract the person representation, $h_p\in\R^d$, from the final-layer hidden state at the closing boundary of \texttt{</person\_records>}. Because this token precedes any prediction questions under causal attention, psychological measurement depends solely on the supplied personal evidence and is invariant to downstream queries.

To map the person representation $h_p$ to continuous scores across all 109 constructs without verbalized text generation, we introduce a \emph{centred cosine readout} with construct-specific scaling and offsets. For each construct $c$, the readout learns a latent direction $v_c\in\R^d$, a positive scale $a_c>0$, and an offset $b_c\in\R$.
With $\phi(x)=x/\lVert x\rVert_2$ denoting $\ell_2$-normalization and $\mu\in\R^d$ denoting an exponential moving average of normalized person representations (updated during training and fixed at inference), the inferred score for construct $c$ is
\begin{equation}\label{eq:readout}
s_c(p)=a_c\,\phi\!\left(\phi(h_p)-\mu\right)^{\!\top}\phi(v_c)+b_c.
\end{equation}
Subtracting the shared centre $\mu$ removes the dominant common direction of language-model hidden states, which typically encodes generic document attributes such as sequence length, formatting, and register, so that $\phi(\phi(h_p)-\mu)$ captures how person $p$ deviates from the average representation. Projecting this centred state onto the unit construct direction $\phi(v_c)$ bounds the directional similarity, while $a_c$ and $b_c$ calibrate each construct readout to the empirical scale of its supervision targets while strictly preserving individual rank orderings.

\subsection{Training framework}\label{sec:method-framework}

Our training framework jointly optimizes three supervision objectives within the shared model, linking heterogeneous empirical evidence to psychological construct scores, individual behavioural decisions, and population response distributions (\cref{fig:overview}).

\subsubsection{Psychological measurement}\label{sec:method-measurement}

We train \chiron{} to infer psychological attributes from personal records using four complementary sources of measurement supervision (\cref{tab:method-data-overview}). Each record $p$ is paired with reference scores $t_{pc}\in[-1,1]$ for its annotated subset of constructs $\mathcal{C}_p$, allowing datasets with partial or non-overlapping annotations to supervise the same 109-construct readout head.

\paragraph{Human-authored text with self-reported assessments.} We pair a person's naturalistic essays, self-reflections, or social-media posts with psychometric scale scores completed by the same individual. Supervising the readout on self-reports grounds the model's latent space in authentic human self-evaluations expressed through unstructured language.

\paragraph{Human questionnaire records with held-out-item targets.} For multi-item human questionnaire datasets, the model receives a subset of a respondent's item answers as input $\mathcal{R}_p$, while the reference score $t_{pc}$ is calculated from a disjoint held-out subset of items using the instrument's official scoring and reverse-keying rules. Predicting construct scores derived from withheld items forces the model to infer underlying latent dispositions across related psychological indicators.

\paragraph{Synthetic questionnaire records with directly calculated scores.} To ensure complete coverage across all 109 constructs, we sample item combinations from each construct's psychometric item pool and simulate internally consistent response profiles across the full trait continuum (rendered either as structured item--response pairs or as naturalized first-person statements; \cref{app:data-construction}). Target scores are computed directly via the instrument's scoring key, teaching the readout exact psychometric item keying and polarity.

\paragraph{Synthetic narratives with comparison-derived scores.} Because human datasets pairing free text with psychometric scores exist for only a fraction of psychological constructs, we extend text supervision across all 109 constructs via calibrated narrative synthesis (\cref{app:data-construction}). A frontier LLM generates 25{,}000 first-person narratives across ten everyday life contexts (e.g., work, family, conflict, leisure) without naming the construct or stating numerical scores. Crucially, generation orthogonally varies construct direction (aligned vs.\ opposing), diagnostic evidence strength (weak, medium, strong), expression mode, and word-count range, explicitly decoupling construct intensity from text length and desirability. An LLM judge then evaluates 500{,}000 pairs of narratives for the same construct (balanced across presentation orders) to determine which expresses the construct definition more strongly. Following ValueFlow \citep{kim2026valueflow}, we fit a Bradley--Terry model to these pairwise comparisons to obtain continuous reference scores $t_{pc}\in[-1,1]$ for 15{,}250 quality-filtered narratives.

Targets from all four sources are normalized to $[-1,1]$. For each record $p$, we minimize the mean squared error over its annotated constructs $\mathcal{C}_p$:
\begin{equation}\label{eq:measurement-loss}
\mathcal{L}_{\mathrm{meas}}
=\frac{1}{|\mathcal{C}_p|}\sum_{c\in\mathcal{C}_p}\bigl(s_c(p)-t_{pc}\bigr)^2.
\end{equation}
Crucially, $\mathcal{L}_{\mathrm{meas}}$ updates both the measurement head parameters $\{v_c,a_c,b_c\}$ and the shared language model adapters, directly shaping the person representation $h_p$ to encode psychometric structure.

\subsubsection{Individual behavioural prediction}\label{sec:method-prediction}

We train \chiron{} to predict how a specific individual will respond to unseen questions or situations given their observed records $\mathcal{R}_p$ (\cref{tab:method-data-overview}). Each training instance pairs visible personal records $\mathcal{R}_p$ with one or more target questions $\mathbf{q}$, supervised by the person's actual withheld answers $\mathbf{y}$ across three data sources (\cref{app:prediction-details}):

\paragraph{Recorded questionnaire answers, ratings, and choices.} Across 19 psychometric and cross-national survey domains, we partition each participant's recorded responses into visible evidence $\mathcal{R}_p$ and held-out target items $(\mathbf{q},\mathbf{y})$. Given the visible answers, the model predicts the participant's exact responses to the withheld items, learning how individual differences co-vary across behaviours and attitudes.

\paragraph{Personal writing with source annotations.} Where datasets pair personal essays or social-media histories with external behavioural or trait annotations, we supply the author's text as $\mathcal{R}_p$ and cast the annotated attributes or scale responses as target questions $(\mathbf{q},\mathbf{y})$.

\paragraph{Behaviours derived from naturalistic text (masked observation prediction).} Most real-world personal writing is unlabelled, creating a data bottleneck for behavioural prediction. We introduce \textit{masked observation prediction}, a self-supervised task formulation that converts unlabelled personal narratives (e.g., from the Blog Authorship Corpus) into structured behavioural prediction instances at scale (\cref{app:prediction-data-construction}). An extraction model first identifies high-confidence, quote-supported first-person observations regarding the author's recurring habits, interests, emotional patterns, and values. We hold out a quarter of an author's extracted observations and transform each withheld observation into a four-option question targeting either a fine-grained behavioural distinction or an abstract dispositional tendency. Providing the remaining observations as $\mathcal{R}_p$ and training \chiron{} to recover the withheld choices teaches the model to perform multi-step behavioural inference from naturalistic personal evidence without human labels.

Across all three prediction sources, we minimize the conditional negative log-likelihood on the reference answer tokens $\mathbf{y}$:
\begin{equation}\label{eq:prediction-loss}
\mathcal{L}_{\mathrm{pred}}
=-\log p_\theta(\mathbf{y}\mid\mathcal{R}_p,\mathbf{q}).
\end{equation}

\subsubsection{Population alignment}\label{sec:method-population}

Population alignment trains \chiron{} to serve as a calibrated simulator of human cohorts in social science experiments. A central barrier in computational social science is that published experimental studies typically report only condition-level summary statistics (such as sample sizes, means, standard deviations, and treatment effect sizes) rather than raw participant-level responses, while raw LLM completions exhibit severe variance compression and effect-size inflation \citep{cui2025largescale,ashokkumar2026experiments,xu2026llmcss}. To overcome this barrier, we reconstruct respondent-level supervision from published aggregate human statistics through \emph{constrained population synthesis}.

We start from a pool of candidate responses generated by a diverse panel of LLMs \citep{cui2025largescale} and synthesize virtual respondent populations whose aggregate statistics match published human findings across training experiments.
Specifically, let $Q$ denote the selected respondents' original answers and $\mathcal{Y}$ their candidate answers after editing. We solve
\begin{equation}\label{eq:population-synthesis}
\widetilde{\mathcal{Y}}
=
\operatorname*{arg\,min}_{\mathcal{Y}\in\mathcal{A}}
D_{\mathrm{source}}(\mathcal{Y},Q)
\quad\text{subject to}\quad
E_{\mathrm{human}}(\mathcal{Y})\leq\varepsilon,
\end{equation}
where $\mathcal{A}$ enforces each condition's exact human sample size $N_c$ and valid item response scales, $E_{\mathrm{human}}(\mathcal{Y})$ measures the normalized discrepancy between the synthetic population's statistics and the published human statistics (computed via each study's exact original scoring and statistical analysis pipeline), and $D_{\mathrm{source}}(\mathcal{Y},Q)$ is the fraction of item responses edited relative to the selected source responses. We solve \cref{eq:population-synthesis} in three optimization stages (\cref{app:synthesis-fit}). Across the 63 training studies, the synthesized populations closely match published human statistics (Cohen's $d$ MAE $=0.012$, $r=0.999$, and distribution NMAE $=0.010$; \cref{tab:synthesis-fit}).

Given each synthesized respondent's answer vector $\widetilde{\mathbf{y}}$ for an experimental condition and its questions $x$, we train the language model via
\begin{equation}\label{eq:population-loss}
\mathcal{L}_{\mathrm{pop}}
=-\log p_\theta(\widetilde{\mathbf{y}}\mid x).
\end{equation}
Crucially, within each experimental condition, all $N_c$ synthesized respondents share the same prompt $x$ but have distinct target answers $\widetilde{\mathbf{y}}_1,\ldots,\widetilde{\mathbf{y}}_{N_c}$ that match the human mean and standard deviation. Minimizing $\mathcal{L}_{\mathrm{pop}}$ across these one-to-many $(\text{prompt } x \to \text{cohort responses})$ pairs trains $p_\theta(\mathbf{y}\mid x)$ to match the empirical cohort distribution. Repeated sampling at inference time can then recover both within-condition human heterogeneity and between-condition treatment effects.

\subsubsection{Joint training recipe}\label{sec:method-recipe}

Our training corpus spans personality, emotional experience, mental health, interpersonal relationships, vocational interests, cultural values, cognitive preferences, and everyday behaviour. It combines questionnaire responses, personal writing, and social-media records with synthetic measurement data and published social science experiments, comprising 53{,}073 measurement records across 109 constructs, 1{,}073{,}073 atomic decisions across 25 behavioural domains, and 23{,}191 virtual respondents across 63 training experiments (\cref{tab:method-data-overview,tab:training-data-detailed}).

We train \chiron{} on a unified mixture of examples from all three objectives using the weighted multi-task loss
\begin{equation}\label{eq:total}
\mathcal{L}_{\mathrm{joint}}
=\mathcal{L}_{\mathrm{pred}}
+\lambda_m\mathcal{L}_{\mathrm{meas}}
+\lambda_p\mathcal{L}_{\mathrm{pop}},
\end{equation}
where all three terms update the shared language model backbone and $\mathcal{L}_{\mathrm{meas}}$ additionally updates the measurement head. We adapt Qwen3.6-35B-A3B \citep{qwen2026qwen36} using LoRA \citep{hu2022lora} and train the adapters and measurement head jointly for two epochs with $\lambda_m=8$ and $\lambda_p=1$ (full optimization hyperparameters in \cref{app:optimization}).

\subsection{Mechanistic and methodological advantages}\label{sec:method-advantages}

\chiron{}'s architecture and training design address fundamental limitations of both general-purpose LLMs and traditional psychological instruments.

\paragraph{Mechanistic advantages over general-purpose LLMs.}
Off-the-shelf frontier LLMs are optimized for helpful, normative conversational assistance instead of scientific measurement or behavioural simulation. This creates three mechanistic failure modes that \chiron{} explicitly resolves:
\begin{enumerate}[leftmargin=1.5em,itemsep=2pt,topsep=2pt]
\item \emph{Direct latent readout avoids the biases of verbalized scoring.} When general LLMs are prompted to score psychological constructs in text, their outputs pass through the autoregressive token generation head, exposing estimates to RLHF-induced social desirability biases that inflate positive traits and compress score dispersion, as well as sensitivity to surrounding irrelevant context and prompt-order recency effects \citep{abdurahman2024perils,ye2025llmpsychometrics,xu2026llmcss}. \chiron{} bypasses text generation entirely by mapping the person representation directly to continuous scores via a dedicated measurement head (\cref{eq:readout}) supervised on psychometric scales and balanced synthetic data. This architectural separation decouples psychological assessment from conversational text generation, avoiding the biases inherent to verbalized scoring.
\item \emph{Joint psychometric supervision structures person representations for behavioural prediction.} Prior approaches either prompt general LLMs with personal context, which relies on surface lexical cues or demographic stereotypes \citep{wang2024misportray,jiang2025indievalue}, or fine-tune black-box predictors without construct grounding, which can learn task-specific shortcuts that generalize poorly \citep{binz2025centaur,xu2026llmcss}. In \chiron{}, the measurement loss $\mathcal{L}_{\mathrm{meas}}$ supervises 109 psychological constructs directly on the shared hidden state $h_p$, and the prediction loss $\mathcal{L}_{\mathrm{pred}}$ conditions on this same state to predict withheld choices $\mathbf{y}$. The measurement training thus regularizes $h_p$ to compress noisy personal records into stable dispositional dimensions that guide downstream prediction, while prediction training ensures that $h_p$ learns decision-relevant individual signals (\cref{fig:prediction-case-study}).
\item \emph{Distributional alignment counteracts mode collapse and restores population dispersion.} Preference optimization of general LLMs concentrates a model's probability mass on modal responses \citep{santurkar2023whose,kirk2024rlhf,murthy2025onefish}.
In population simulation, this suppresses within-condition standard deviations, inflates standardized effect sizes, and produces false positives on null findings \citep{cui2025largescale,ashokkumar2026experiments}.
\chiron{} reverses this effect through constrained population synthesis and distributional alignment (\cref{eq:population-synthesis,eq:population-loss}). Training across the full cohort penalizes over-concentrated outputs and restores calibrated population distribution.
\end{enumerate}

\paragraph{Methodological advantages over traditional psychological tools.}
Compared with conventional psychometric questionnaires, laboratory experiments, and narrow text-analysis tools (such as lexicon dictionaries or single-trait classifiers), \chiron{} offers three distinct methodological advantages for behavioural science:
\begin{enumerate}[leftmargin=1.5em,itemsep=2pt,topsep=2pt]
\item \emph{Unified, non-invasive assessment operates across heterogeneous evidence.} Traditional psychometric scales require participants to complete fixed, structured item batteries, limiting sample scale and precluding retrospective measurement from naturalistic digital traces, historical texts, or qualitative interviews. Conversely, lexicon tools (e.g., word-count dictionaries) and task-specific classifiers are restricted to narrow trait sets and fail to capture compositional meaning or context. \chiron{} infers 109 standardized, theory-grounded constructs simultaneously from arbitrary mixtures of unstructured text, behavioural records, and questionnaire response.
\item \emph{Shared representations bridge construct interpretability and behavioural predictability.} Psychological research has long faced a divide between explanatory construct measurement (which yields interpretable trait profiles but often weakly predicts specific situational choices) and black-box behavioural prediction \citep{hofman2021integrating,xu2026llmcss}. By coupling an explicit 109-construct measurement head with a generative behavioural predictor on the same person representation, \chiron{} provides both transparent psychological profiles and accurate predictions of individual behaviour.
\item \emph{Calibrated population simulation enables in silico pretesting of experimental designs.} Conducting human-subject experiments across many candidate framing conditions, stimuli, or subpopulations is resource-intensive. While uncalibrated LLMs distort effect magnitudes, \chiron{}'s population alignment enables researchers to pretest experimental protocols, estimate plausible effect sizes and response distributions, and prioritize high-potential hypotheses in silico before deploying confirmatory human studies.
\end{enumerate}

\section{Experiments}\label{sec:experiments}

We evaluate \chiron{} through psychological measurement, individual behavioural prediction, and population simulation. 
Measurement examines agreement with human assessments, alignment with literature-derived construct relationships, robustness to irrelevant changes, and responsiveness to experimental manipulations of construct-relevant evidence. Prediction tests responses withheld from personal records, while population simulation compares generated response distributions and experimental effects with human findings. All main evaluations use the same jointly trained model.
Across all three evaluation settings, we compare \chiron{} with three frontier models (GPT-6 Sol, Claude Opus 5.5 and DeepSeek-V4.1-Flash) and the unadapted Qwen3.6-35B-A3B base model.

\subsection{Psychological measurement}\label{sec:exp-humangold}

\begin{figure}[!pt]
\centering
\makebox[\linewidth][c]{\begin{minipage}{180mm}
\includegraphics[width=\linewidth]{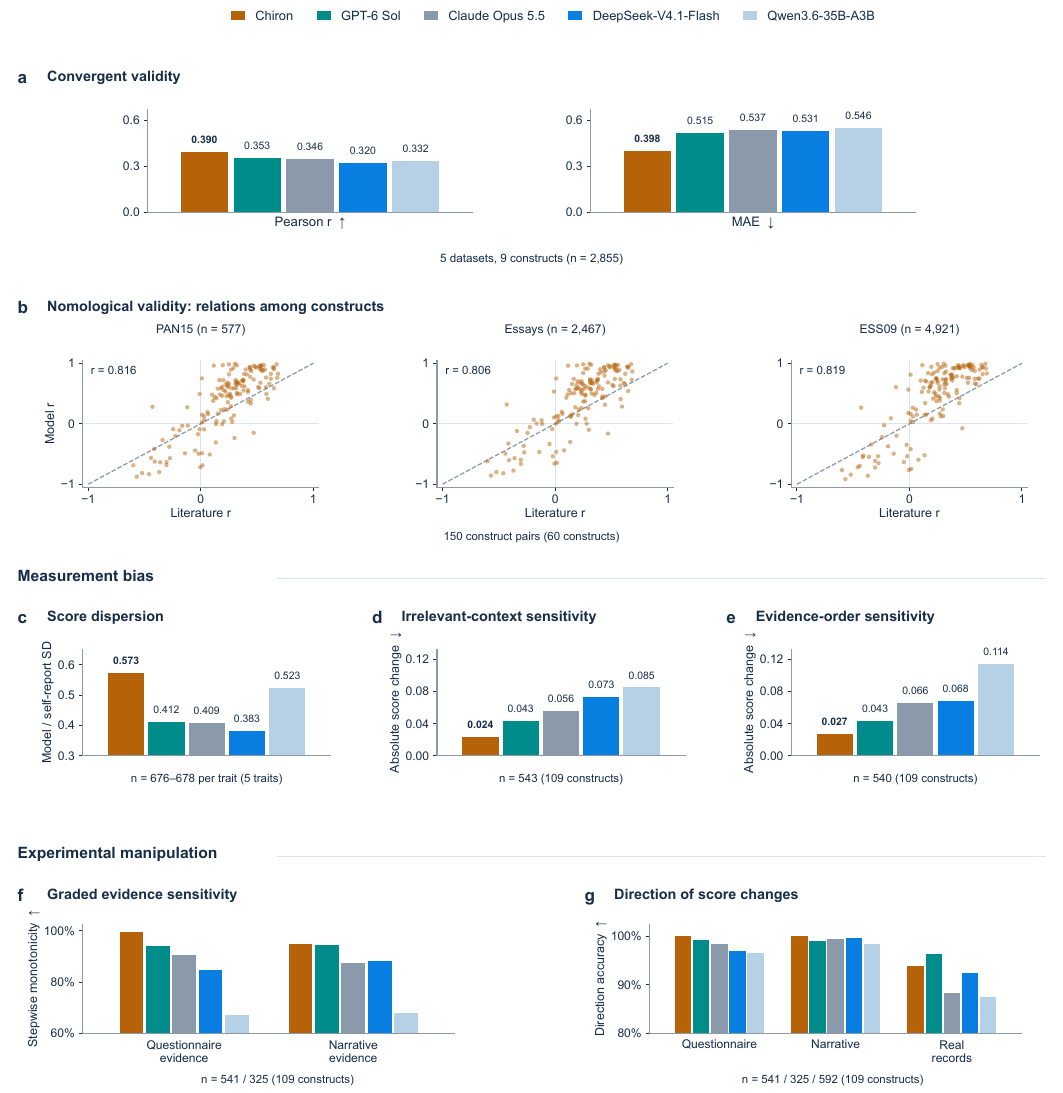}
\caption{Evaluation results of psychological measurement. \textbf{a}, Pearson $r$ and MAE against self-reports (five datasets, nine constructs). \textbf{b}, Construct correlations versus literature estimates (150 pairs, 60 constructs); dashed lines indicate $y=x$.
\textbf{c}, Standard deviation ratio of model scores to self-reports across five personality traits on PANDORA; 1 indicates equal dispersion.
 \textbf{d}, Absolute score changes under irrelevant-context perturbations (109 constructs). \textbf{e}, Absolute score changes under evidence-order perturbations (109 constructs). 
 \textbf{f}, Stepwise monotonicity across four graded evidence levels (109 constructs). 
 \textbf{g}, Direction accuracy for low--high evidence interventions on questionnaires, narratives and real records.}
\label{fig:measurement-validity}
\end{minipage}}
\end{figure}

\subsubsection{Convergent validity}

Convergent validity measures the agreement between different assessments of the same psychological construct. We compare scores inferred from personal text with self-report assessments completed by the same authors, using Pearson correlation and mean absolute error (MAE). The evaluation spans five datasets (COVID-19 RW3D, Buechel empathy, PANDORA, PsihoRo and Student-SADD) covering nine constructs. 
Presented in \cref{fig:measurement-validity}a, we calculate Pearson correlation and MAE within each dataset and report the unweighted mean across datasets. 
For GPT-6 Sol, 127 requests that were refused after retries are excluded when computing its correlation and MAE.

\paragraph{\chiron{} infers psychological constructs from personal text more accurately than frontier models.}
Across the five datasets, \chiron{} achieves a mean Pearson $r$ of 0.390 and MAE of 0.398, outperforming all four baselines ($r=0.320\text{--}0.353$ and MAE $=0.515\text{--}0.546$). Detailed results by dataset and construct are presented in \cref{tab:measurement-detailed-common}.

\subsubsection{Nomological validity}\label{sec:exp-nomological}

Nomological validity assesses whether relationships among measured constructs agree with established psychological theory and empirical findings \citep{cronbach1955construct}. We examine this by calculating correlations among \chiron{}'s construct scores across individuals and comparing them against a literature benchmark of median empirical correlations compiled from published psychometric studies. The evaluation covers 150 relationships among 60 constructs, assessed separately across three datasets with distinct record types: PAN15, Essays, and ESS09. The results are plotted in \cref{fig:measurement-validity}b. Benchmark construction and evaluation details provided in \cref{app:eval-kg}.

\paragraph{\chiron{} recovers empirical inter-construct relationships reported in the psychological literature.}
Across the three datasets, correlations among \chiron{}'s inferred scores align closely with literature benchmarks, reaching Pearson correlations of 0.816 on PAN15, 0.806 on Essays and 0.819 on ESS09, with sign agreement in 91.3--93.3\% of construct pairs (\cref{tab:kg-alignment}). This consistency across social-media posts, personal essays, and survey responses indicates that \chiron{} preserves a coherent nomological structure across different forms of personal evidence.

\subsubsection{Measurement bias}\label{sec:exp-robustness}

Measurement bias refers to systematic distortions in inferred scores that are unrelated to the target construct. We examine three types of bias across \cref{fig:measurement-validity}c--e:
(1) \emph{score compression} (\cref{fig:measurement-validity}c), whether inferred scores across individuals are compressed into a narrower range than human self-reports, measured by the ratio of model-score standard deviation to self-report standard deviation averaged across the Big Five traits on PANDORA (where 1 indicates equal dispersion);
(2) \emph{irrelevant-context sensitivity} (\cref{fig:measurement-validity}d), whether scores drift when construct-irrelevant background statements in the input are replaced while the target evidence is held fixed, measured by the mean absolute score change ($n=543$ units across all 109 constructs); and
(3) \emph{evidence-order sensitivity} (\cref{fig:measurement-validity}e), whether scores depend on record ordering, measured by the mean absolute score change when the presentation order of statements in an input is reversed ($n=540$ units across all 109 constructs).
Evaluation protocols and detailed results are provided in \cref{app:measurement-bias}.

\paragraph{\chiron{} exhibits less score compression and greater invariance to irrelevant context and evidence ordering than baseline models.}
On PANDORA (\cref{fig:measurement-validity}c), all models are compressed relative to self-reports, but \chiron{} reaches a mean standard-deviation ratio of 0.573, remaining closer to human dispersion than the three frontier models (0.383--0.412) and the base model (0.523). The frontier models' stronger compression relative to both \chiron{} and the base model corroborates the mechanism discussed in \cref{sec:method-advantages}: post-training and preference alignment homogenize verbalized assessments toward modal values \citep{kirk2024rlhf,murthy2025onefish,xu2026llmcss}, whereas \chiron{}'s centred representation readout better preserves inter-individual dispersion. Under input perturbations (\cref{fig:measurement-validity}d,e), \chiron{} likewise achieves the smallest mean absolute score changes: 0.0238 when replacing construct-irrelevant context (versus 0.0433--0.0850 across baselines, reducing context-induced drift by 45.0--71.9\%) and 0.0270 when reversing evidence order (versus 0.0433--0.1139 across baselines).

\subsubsection{Experimental manipulation}

In psychometric validation, a valid measurement instrument must respond predictably to experimental manipulations of construct-relevant evidence while holding background context constant. We conduct controlled interventions that systematically manipulate the diagnostic evidence for a target construct across all 109 constructs: we progressively replace low-level evidence with high-level evidence across four graded levels in questionnaire and narrative inputs, and insert contrasting low- or high-level narratives into otherwise unchanged real personal records. We evaluate stepwise monotonicity as the proportion of adjacent transitions across the four graded manipulation levels in which the score strictly increases, and direction accuracy as the proportion of cases in which the high-evidence manipulation receives a higher score than the low-evidence manipulation. The results are shown in \cref{fig:measurement-validity}f,g; experimental setups and detailed results are provided in \cref{app:measurement-interventions}.

\paragraph{\chiron{} responds monotonically and in the expected direction to experimental manipulations of construct-relevant evidence.}
Across all 109 constructs, \chiron{}'s scores strictly increase across 99.4\% of adjacent steps for questionnaire evidence and 94.7\% for narrative evidence, compared with 67.2--94.0\% and 68.0--94.3\% for the four baselines. For direction accuracy, \chiron{} reaches 100.0\% on both questionnaire and narrative manipulations (versus 96.5--99.3\% and 98.5--99.7\% for the baselines) and 93.8\% when contrasting narratives are inserted into real personal records (versus 87.6--96.4\% for the baselines).

\begin{figure}[!pt]
\centering
\makebox[\linewidth][c]{%
\begin{minipage}{180mm}
\centering
\includegraphics[width=\linewidth]{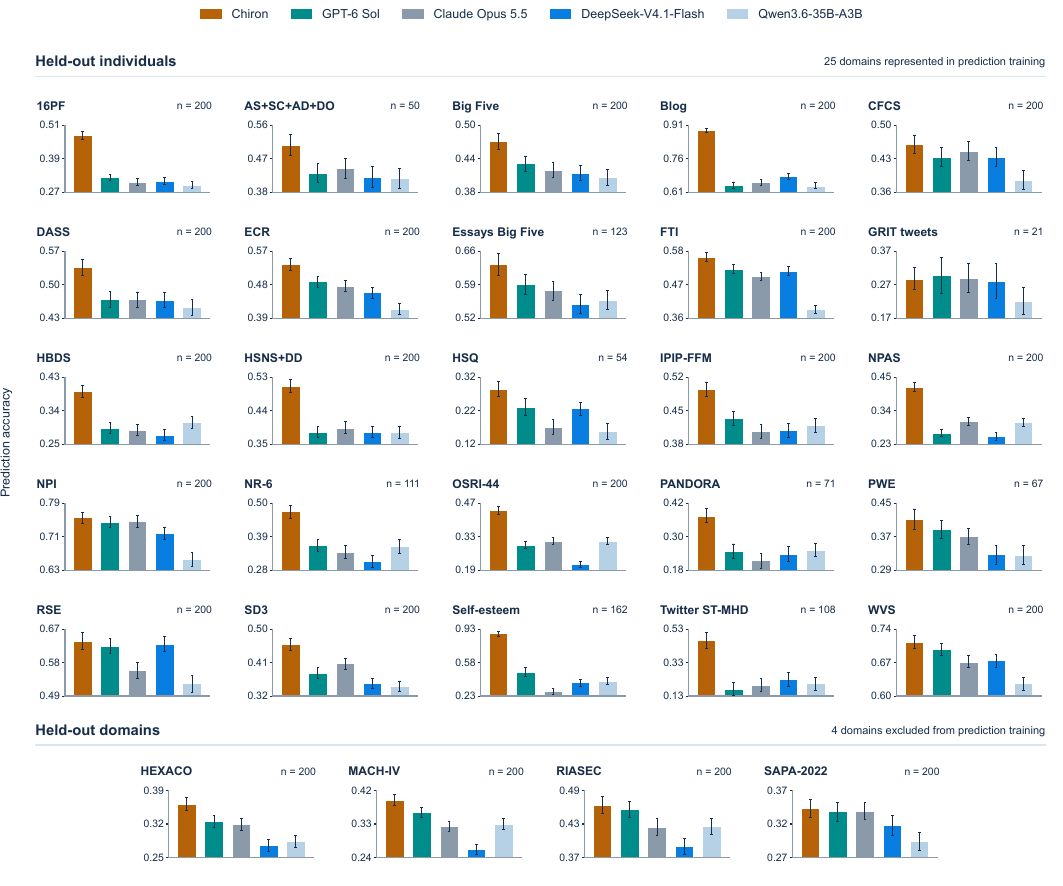}
\caption{Prediction accuracy across 29 behavioural domains. Bars show mean prediction accuracy ($\pm1$ SEM) across test cases within each domain.}
\label{fig:prediction-domains}
\end{minipage}%
}
\end{figure}

\subsection{Individual behavioural prediction}\label{sec:exp-prediction}

We evaluate whether \chiron{} can predict a person's withheld responses, choices, or behavioural attributes, based on their available personal records. The benchmark spans 29 behavioural domains and examines two generalization settings:
(1) \emph{held-out individuals}, which tests unseen participants (or source records) across the 25 domains included in behavioural prediction training;
and (2) \emph{held-out domains}, which tests four additional domains excluded from behavioural prediction training (HEXACO, MACH-IV, RIASEC and SAPA-2022).
All models receive the same observed records and target questions. The test set contains 4{,}767 cases: 3{,}967 cases in the 25 represented domains and 800 cases in the four held-out domains. We compute prediction accuracy for each test case (the proportion of withheld target questions answered correctly for that person) and report the mean and standard error of the mean (SEM) across cases within each domain. For GPT-6 Sol, 60 requests that were refused after retries are excluded when computing its mean accuracy, whereas malformed outputs (two Blog cases for Qwen3.6-35B-A3B) are scored as zero. Domain-level results across both generalization settings are presented in \cref{fig:prediction-domains}, with further details provided in \cref{app:prediction-details}. 

\paragraph{\chiron{} more accurately predicts individual behavior, generalizing to both unseen individuals and unseen behavioural domains.}
Across all 29 domains, \chiron{} reaches a mean prediction accuracy of 0.504 (averaging equally across domains), compared with 0.382--0.418 for the four baselines, and ranks first in 28 of the 29 domains. 
This advantage holds across both generalization levels: on held-out individuals within the 25 represented domains, \chiron{} attains a mean prediction accuracy of 0.522 (versus 0.389--0.425 for the baselines), and on the four held-out domains excluded from prediction training, it remains the top-performing model on every domain with a mean prediction accuracy of 0.391 (versus 0.312--0.370).

\begin{figure}[!thp]
\centering
\includegraphics[width=0.96\linewidth]{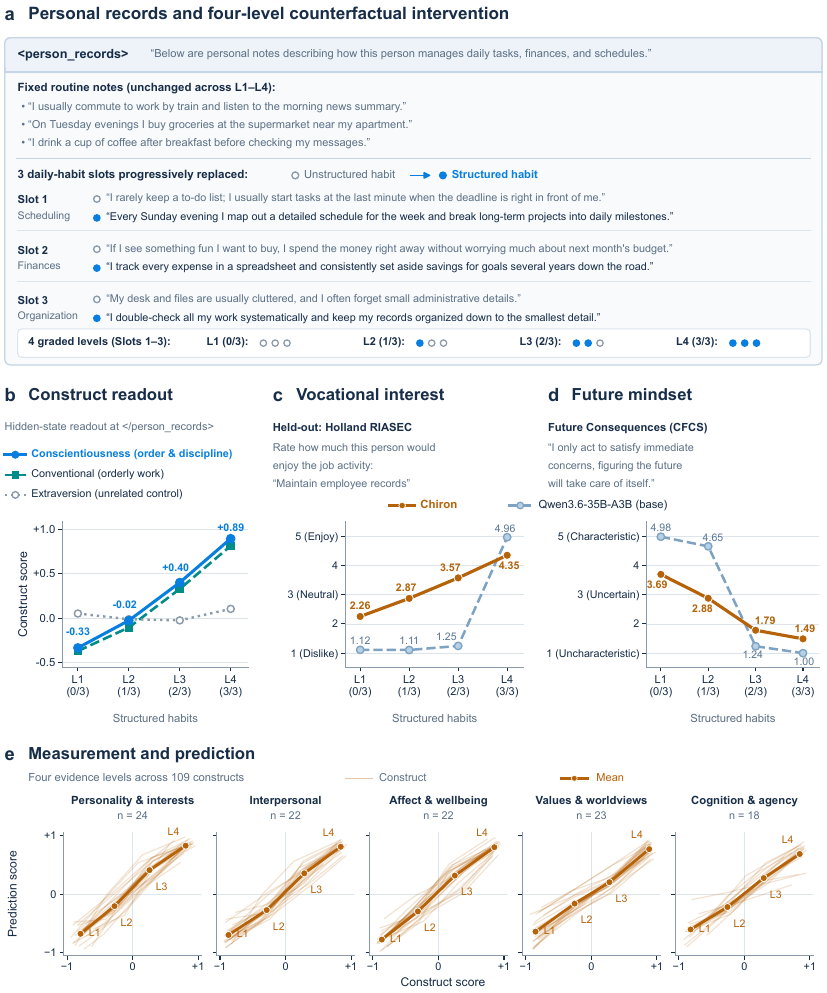}
\caption{Linking latent construct measurement to behavioural prediction under graded evidence interventions. \textbf{a}, Counterfactual intervention on personal records: three fixed routine notes are paired with three daily-habit slots (scheduling, finances, and organization) that are progressively replaced from unstructured (L1, $0/3$) to structured (L4, $3/3$). \textbf{b}, \chiron{} measurement results across L1--L4. \textbf{c,d}, Expected item ratings across L1--L4 for \chiron{} and the base model on a held-out vocational interest item (Holland RIASEC; \textbf{c}) and a reverse-keyed future-orientation item (CFCS; \textbf{d}). \textbf{e}, measurement scores ($x$-axis) and target-item prediction score ($y$-axis) across four graded evidence levels (L1--L4) for all 109 constructs ($n$ denotes constructs per category; 2{,}180 inputs and 6{,}540 item predictions). Prediction scores average expected responses across three target items per construct, keyed toward the high pole and rescaled to $[-1,1]$. Thin lines trace individual constructs (averaged over five synthetic backgrounds); thick lines indicate category means.}
\label{fig:prediction-case-study}
\end{figure}

To examine how psychometric supervision and latent construct measurement support downstream behavioural prediction (\cref{sec:method-advantages}), we trace \chiron{}'s construct readouts at \texttt{</person\_records>} and its item-level predictions across four graded evidence levels (L1--L4) in two settings (\cref{fig:prediction-case-study}).

(1) We conduct a \emph{cross-domain counterfactual case study} on naturalistic personal records (\cref{fig:prediction-case-study}a--d). Here, we hold three everyday routine notes fixed and progressively replace three unstructured habits in task scheduling, personal finances, and work organization with structured alternatives from L1 ($0/3$ structured) to L4 ($3/3$ structured) (\cref{fig:prediction-case-study}a). 
Conditioning on the person representation at \texttt{</person\_records>}, we read out \chiron{}'s construct scores (\cref{fig:prediction-case-study}b) and compare expected, predicted option ratings $E[y]=\sum_{k=1}^{5}k\,p(y=k\mid\mathcal{R}_p,q)$ between \chiron{} and the unadapted Qwen3.6-35B-A3B base model. Two models predict item response on a held-out vocational domain (Holland's RIASEC inventory, rating job activity \emph{``Maintain employee records''} from 1 = Dislike to 5 = Enjoy; \cref{fig:prediction-case-study}c) and a reverse-keyed future-orientation item (Consideration of Future Consequences Scale, CFCS, \emph{``I only act to satisfy immediate concerns, figuring the future will take care of itself''}; \cref{fig:prediction-case-study}d).

(2) We extend the case study to a \emph{systematic within-instrument test} across all 109 constructs (\cref{fig:prediction-case-study}e). For each construct, we pair each of five synthetic backgrounds with three item responses. Keeping the background fixed, we change these responses one at a time from the lowest- to the highest-scoring answer, creating four evidence levels (L1--L4).
For each input, \chiron{} measures the construct and predicts responses to three other target items. The five backgrounds and four levels yield 2{,}180 inputs and 6{,}540 item predictions across all constructs.
Experimental setups and evaluation details are provided in \cref{app:measurement-prediction-linkage}.

\paragraph{Construct measurement and cross-domain prediction respond consistently to graded evidence.}
In the counterfactual case study, \chiron{}'s inferred conscientiousness increases monotonically across levels ($-0.33 \to -0.02 \to +0.40 \to +0.89$) alongside Holland's \emph{conventional} vocational interest, which captures preference for orderly, rule-governed clerical work ($-0.37 \to -0.10 \to +0.33 \to +0.81$). As a comparison, an unrelated trait (extraversion) remains flat near zero ($-0.02$ to $+0.11$). Given the same personal records, \chiron{}'s expected option ratings shift smoothly and monotonically on both the held-out RIASEC item ($2.26 \to 2.87 \to 3.57 \to 4.35$; \cref{fig:prediction-case-study}c) and the reverse-keyed CFCS item ($3.69 \to 2.88 \to 1.79 \to 1.49$; \cref{fig:prediction-case-study}d). 
By contrast, the unadapted Qwen3.6-35B-A3B base model fails to integrate graded behavioural evidence and exhibits a pronounced extreme-option selection bias characteristic of general-purpose LLMs on Likert-type scales. Instead of calibrating intermediate ratings to mixed evidence, it collapses probability mass onto the scale endpoints (1 or 5), remaining near the floor on the held-out RIASEC item across L1--L3 ($1.12 \to 1.11 \to 1.25$) before jumping abruptly to $4.96$ at L4, and switching sharply between opposite extremes on the CFCS item ($4.98 \to 4.65$ versus $1.24 \to 1.00$).

\paragraph{Latent construct scores summarize the graded evidence behind within-instrument prediction.}
Across all 109 constructs in the systematic within-instrument test (\cref{fig:prediction-case-study}e), \chiron{}'s construct scores and mean prediction scores increase together in 99.27\% of adjacent transitions. To examine whether the construct score accounts for the prediction itself, we treat each construct's 20 records, formed by crossing its five backgrounds with the four evidence levels, as separate observations. Within each construct, we first regress the mean prediction score on the construct score alone, and then repeat the regression after adding the evidence level, coded as 0, 1, 2 and 3. The construct score linearly explains the mean prediction score, with a mean Pearson $r$ of 0.970 ($R^2=0.942$) across constructs, and adding the evidence level increases $R^2$ by only 0.007. The continuous readout therefore already summarizes the graded evidence on which the prediction depends.

\begin{figure}[!thbp]
    \centering
    \makebox[\linewidth][c]{\begin{minipage}{180mm}
    \includegraphics[width=\linewidth]{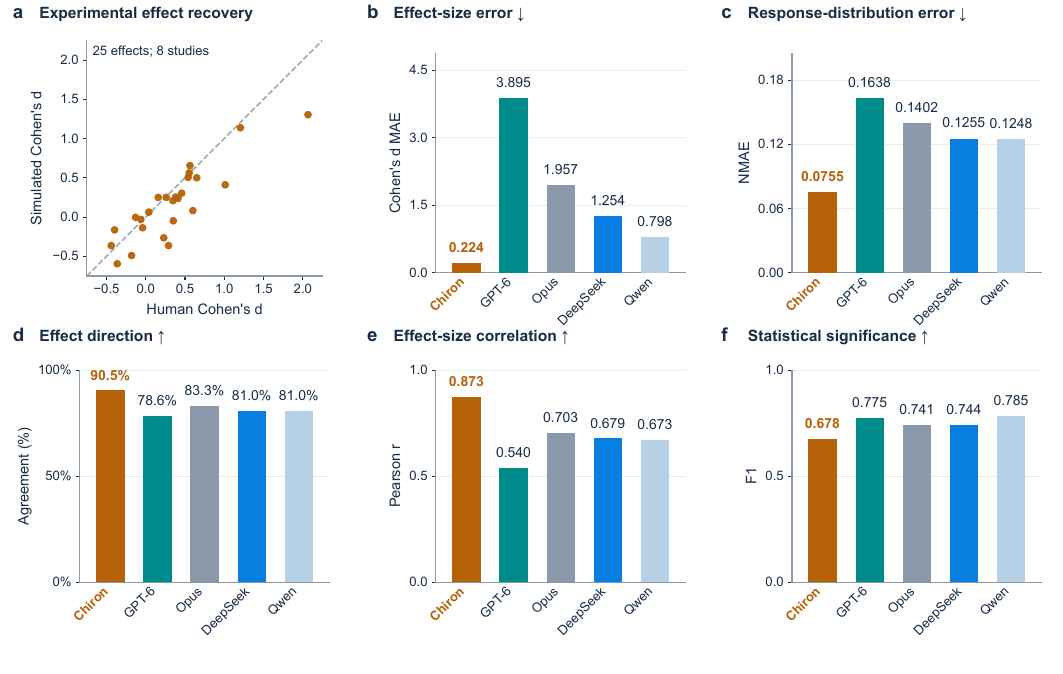}
    \caption{Population simulation across 15 held-out studies. \textbf{a}, Simulated versus human Cohen's $d$ across 25 eligible effects from eight studies; the dashed line indicates $y=x$. \textbf{b,c}, Effect-size estimation error (Cohen's $d$ MAE, $n=25$) and response-distribution error (NMAE, $n=34$). \textbf{d--f}, Effect-direction accuracy ($n=42$), effect-size Pearson correlation ($n=25$), and statistical significance F1 ($n=53$). Arrows indicate the preferred direction for each metric.}
    \label{fig:population-results}\label{fig:results}
    \end{minipage}}
\end{figure}

\subsection{Population simulation}\label{sec:exp-population}

We evaluate whether simulated populations reproduce human response distributions and experimental effects across social science experiments. The benchmark splits studies into 63 training studies and 15 held-out test studies; all conditions from a study remain in the same split, and only training studies contribute to population synthesis and joint training. Across the 15 test studies, each model generates 4{,}277 planned virtual participants over 45 experimental conditions, following the replication protocol of \citet{cui2025largescale}. 
We apply each study's original scoring procedure to compare simulated and human outcomes in terms of effect magnitudes (Cohen's $d$ MAE and Pearson correlation), effect direction accuracy, statistical significance F1, and normalized mean absolute error (NMAE) of response distributions. For Claude Opus 5.5, 98 refused requests are not imputed. GPT-6 Sol, Claude Opus 5.5 and DeepSeek-V4.1-Flash each have one zero-variance condition pair in AMJ\_4\_2. Results across the 15 held-out studies are presented in \cref{fig:population-results}, with further details provided in \cref{app:population-details}.

\paragraph{\chiron{} more faithfully reproduces human experimental effects and response distributions across held-out studies than frontier models.}
Across the 15 held-out studies, \chiron{} substantially outperforms all four baselines on four of the five evaluation metrics. For treatment effect estimation, \chiron{} reduces Cohen's $d$ MAE to 0.224 (compared with 0.798--3.895 for the baselines) and reaches an effect-size correlation of $r=0.873$ (versus 0.540--0.703), while achieving the highest effect direction accuracy at 90.5\% (versus 78.6--83.3\%). It also matches human response distributions more closely, with a distribution NMAE of 0.0755 compared with 0.1248--0.1638 for the baselines. 

Notably, stronger frontier models exhibit larger distributional and effect-size distortions than the unadapted base model: Cohen's $d$ MAE increases monotonically from 0.798 for Qwen3.6-35B-A3B to 1.254--3.895 across the three frontier models, accompanied by a parallel rise in distribution NMAE (with the three frontier models even collapsing to zero within-condition variance in AMJ\_4\_2). This pattern directly reflects the mechanism analysed in \cref{sec:intro,sec:method-advantages}: standard instruction tuning and preference alignment suppress within-condition response variance and homogenize simulated participants, which systematically inflates standardized effect sizes ($d$, where between-condition mean differences are divided by pooled within-condition standard deviation). 
This mechanism also explains why \chiron{} obtains a lower statistical significance F1 than the baselines (0.678 versus 0.741--0.785 across the 53 tests). Because published experimental benchmarks consist predominantly of statistically significant findings ($p<0.05$) and the F1 metric rewards true positives while ignoring true negatives, models that indiscriminately produce small $p$-values are systematically favoured. By underestimating within-condition variance and exaggerating treatment effects, the four baselines declare 19--22 more comparisons statistically significant than \chiron{}, raising true positives from 20 to 29--31 but correctly identifying only 2--5 true negatives on null human effects (compared with 14 for \chiron{}). In contrast, \chiron{} preserves realistic response dispersion and produces more conservative, better-calibrated effect magnitudes.

\section{Conclusion and Discussion}\label{sec:discussion}

We introduced \chiron{}, a foundation model to measure and predict human behaviour that jointly learns psychological measurement, individual behavioural prediction, and population alignment within a shared language model. Combining psychometric supervision, self-supervised masked observation prediction, and constrained population synthesis enables scalable learning from heterogeneous individual records and aggregate study statistics. Across our evaluations, \chiron{} produces more accurate and psychometrically valid construct measurements than frontier LLMs, recovering literature-derived nomological networks while mitigating score compression, irrelevant-context drift, and evidence-order sensitivity. For individual prediction, \chiron{} outperforms frontier models across 29 behavioural domains (up to 28.4\% relative improvement), generalizing to both unseen individuals and held-out domains. In population simulations across 15 held-out social science experiments, it counteracts the variance collapse of aligned frontier LLMs and reduces effect-size estimation error by 82.2--94.3\%. Together, these findings establish \chiron{} as a unified computational instrument for measuring psychological attributes and predicting human behaviour across individual and population scales.

\paragraph{Towards addressing fundamental challenges for LLMs in computational social science.}
In \cref{sec:intro}, we outlined four methodological challenges that limit general-purpose LLMs in psychology and computational social science. Our empirical findings show promise for addressing each challenge:
\begin{itemize}[leftmargin=1.5em,itemsep=2pt,topsep=2pt]
\item \emph{Decoupled latent readouts mitigate psychometric biases in unstructured assessment.} Off-the-shelf LLMs suffer from social-desirability inflation, variance compression, and framing sensitivity when verbalizing psychological measurement through text generation \citep{abdurahman2024perils,ye2025llmpsychometrics,brickman2025assessment}. \chiron{}' architecture measures constructs directly from person representations, bypassing token generation. This design yields closer agreement with human self-reports, alignment with literature-derived construct relationships, fuller inter-individual score dispersion, smaller drift under irrelevant context and evidence-order changes, and monotonic responses to experimental manipulations of construct-relevant evidence (\cref{fig:measurement-validity}).
\item \emph{Jointly learning construct readouts and behavioural prediction reconciles interpretability with predictability.} Bridging explanatory psychological constructs and out-of-sample behavioural prediction has been a longstanding challenge \citep{hofman2021integrating}. Joint supervision on a shared person representation yields transparent trait profiles and accurate predictions of withheld behavior (\cref{fig:prediction-domains}). 
Trait scores and predicted item answers shift together when trait-related evidence strengthens (\cref{fig:prediction-case-study}), suggesting measurement and prediction draw on the same information. The same person representation thus links interpretable traits to the choices the model predicts.
\item \emph{Distributional alignment mitigates homogenization and variance collapse in simulated populations.} Heavily posttrained frontier LLMs often collapse within-condition variance and exaggerate standardized treatment effects in experimental simulation \citep{kozlowski2025simulating,ashokkumar2026experiments,cui2025largescale}. Constrained population synthesis trains \chiron{} to match human statistics. It produces more realistic response distributions, better-calibrated effect magnitudes, and fewer false positives on null human findings (\cref{fig:population-results}). These results point toward the potential of using \chiron{}-driven simulations to pretest experimental designs, screen survey instruments, and gauge plausible effect sizes before confirmatory human studies.
\item \emph{Self-supervision and synthetic data mitigate privacy and data-access constraints.} Our results suggest that \chiron{} can be trained and calibrated without raw participant-level experimental releases or exhaustive psychometric labels on every record \citep{xu2026llmcss}. For measurement training, psychometrically labelled human text is scarce. We generate first-person stories for all 109 constructs and fits measurement supervision from pairwise-calibrated construct scores. 
For prediction training, masked observation prediction allows learning individual behaviour from unlabelled personal writing.
For distributional alignment, constrained population synthesis fits virtual respondents to published human statistics.
\end{itemize}

\paragraph{Limitations and future directions.}
We note several methodological scope boundaries and directions for extension. First, our measurement head uses a fixed inventory of 109 constructs instead of open-ended trait descriptions. These 109 constructs are curated from established psychometric taxonomies (\cref{tab:constructs}) to provide broad, non-redundant coverage across five core domains of human variation: personality and vocational interests, self-evaluation and interpersonal functioning, affect and mental health, moral and cultural values, and cognitive self-regulation. This inventory offers sufficient theoretical completeness to represent individual differences. Second, our behavioural prediction objective targets structured choices, ratings, and classifications rather than free-text generation. Focusing on structured decisions removes surface linguistic confounds, enables exact evaluation against human ground truth, and matches standard survey and experimental formats. Extending \chiron{} to open-ended dialogue and free-form behavioural generation is an important next step. Third, our model encodes a person's records as a static snapshot. Incorporating longitudinal timestamps will help separate stable traits from transient states over time.

\paragraph{Ethical considerations.}
Models that infer psychological attributes and predict human behaviour carry dual-use risks that require clear boundaries. First, psychological inference from digital traces must respect informed consent and autonomy. \chiron{} is built for scientific research only, and must not be used for unauthorized profiling, covert surveillance, or high-stakes automated decisions in hiring, credit, insurance, or clinical diagnosis. Second, because inferring individual dispositions and framing effects could enable targeted persuasion or manipulation, personalized AI systems using psychometric representations require human oversight and safeguards against manipulative steering. Third, in population simulation, calibrated in silico cohorts are designed for hypothesis generation and experimental pretesting. Validation with real human subjects remains essential before drawing scientific conclusions.

\bibliography{references}

\clearpage

\appendix

\vbox{%
  \hsize\textwidth
  \linewidth\hsize
  \vskip 0.1in
  \hrule height 4pt
  \vskip 0.25in
  \vskip -\parskip
  \centering
  {\LARGE\bf A Foundation Model\\to Measure and Predict Human Behaviour\\[0.15cm](Appendix)}
  \vskip 0.29in
  \vskip -\parskip
  \hrule height 1pt
  \vskip 0.09in
}

\startcontents[appendix]
\vspace{0.5cm}
\printcontents[appendix]{}{1}{\setcounter{tocdepth}{2}}
\vspace{1cm}

\clearpage

\section{Data and psychological constructs}\label{app:data}

\subsection{Data overview}\label{app:data-overview}

\Cref{tab:training-data-detailed} summarizes the training and evaluation data sources, psychological constructs, data formats, sample counts, and split units across psychological measurement (M), individual behavioural prediction (P), and population simulation (Pop). Use of each dataset remains subject to its upstream licence and access terms. Within each individual-level training objective, the evaluation datasets are disjoint from training datasets based on participant identity where available and by source record otherwise.

{
\makeatletter
\@ifundefined{flrow@oriLT@array}{}{%
  \let\LT@array\flrow@oriLT@array
  \let\longtable\FB@ori@longtable
  \let\endlongtable\FB@ori@endlongtable
  \let\LT@output\FB@ori@LT@output
}
\long\def\LT@makecaption#1#2#3{%
  \LT@mcol\LT@cols c{\hbox to\z@{\hss\parbox[t]\LTcapwidth{%
    \reset@font
    \sbox\@tempboxa{#1{#2: }#3}%
    \ifdim\wd\@tempboxa>\hsize
      #1{#2: }#3%
    \else
      \hbox to\hsize{\hfil\box\@tempboxa\hfil}%
    \fi
    \endgraf\vskip\baselineskip}%
  \hss}}}
\makeatother
\setlength{\LTleft}{0pt}
\setlength{\LTright}{0pt}
\setlength{\LTpre}{8pt}
\setlength{\LTpost}{10pt}
\setlength{\LTcapwidth}{\linewidth}
\begin{longtable}{@{}
>{\raggedright\arraybackslash}p{\dimexpr0.20\linewidth-2\tabcolsep\relax}
>{\raggedright\arraybackslash}p{\dimexpr0.23\linewidth-2.3\tabcolsep\relax}
>{\raggedright\arraybackslash}p{\dimexpr0.20\linewidth-2\tabcolsep\relax}
>{\raggedright\arraybackslash}p{\dimexpr0.13\linewidth-1.3\tabcolsep\relax}
>{\raggedright\arraybackslash}p{\dimexpr0.11\linewidth-1.1\tabcolsep\relax}
>{\raggedright\arraybackslash}p{\dimexpr0.13\linewidth-1.3\tabcolsep\relax}@{}}
\caption{Training and evaluation data across domains and sources. M, psychological measurement; P, individual behavioural prediction; Pop, population alignment and simulation. Population counts additionally show the number of studies in parentheses.}\label{tab:training-data-detailed}\\
\toprule
\textbf{Domain / source} & \textbf{Psychological construct} & \textbf{Data format} & \textbf{Training records} & \textbf{Eval. records} & \textbf{Eval. split} \\
\midrule
\endfirsthead
\multicolumn{6}{@{}l}{\emph{Table \ref{tab:training-data-detailed} continued}}\\
\toprule
\textbf{Domain / source} & \textbf{Psychological construct} & \textbf{Data format} & \textbf{Training records} & \textbf{Eval. records} & \textbf{Eval. split} \\
\midrule
\endhead
\midrule
\multicolumn{6}{r@{}}{\emph{continued on next page}}\\
\endfoot
\bottomrule
\endlastfoot
\multicolumn{6}{@{}l}{\textbf{Personal writing and social-media records}}\\*
\midrule
\noalign{\nobreak}
COVID-19 RW3D \citep{vandervegt2023rw3d} & Anger and anxiety & Personal writing with self-reported scores & M:~1{,}357 & M:~1{,}133 & Participants \\
Buechel empathy \citep{buechel2018empathy} & Empathic concern & Written reactions with self-reported scores & M:~1{,}009 & M:~842 & Participants \\
myPersonality \citep{kosinski2013private} & Big Five traits & Social-media writing with assessment scores & M:~56 & --- & Participants \\
PAN15 Author Profiling \citep{rangel2015pan15} & All 109 constructs (nomological validity) & Social-media writing & --- & M:~577 & Evaluation only \\
PANDORA \citep{gjurkovic2021pandora} & Big Five traits & Reddit records with scores and trait categories & M:~854\newline P:~1{,}275 & M:~693\newline P:~71 & Authors within each objective \\
PsihoRo \citep{ciobotaru2026psihoro} & Depression and anxiety & Personal writing with assessment scores & M:~109 & M:~92 & Participants \\
Student-SADD \citep{tlachac2022studentsadd} & Depression & Student writing with assessment scores & M:~114 & M:~95 & Participants \\
Blog Authorship Corpus \citep{schler2006blog} & Experiences, preferences and behaviour & Blog-derived observations and masked-observation questions & P:~4{,}366 & P:~200 & Authors \\
Essays Big Five \citep{pennebaker1999linguistic} & Big Five traits & Personal essays with trait labels & P:~2{,}221 & M:~2{,}467\newline P:~123 & Source records \\
GRIT Tweets \citep{maheshwari2019nitty} & Perseverance and consistency of interests & Tweets and profiles paired with grit-scale answers & P:~1{,}850 & P:~21 & Authors \\
LoST self-esteem \citep{garg2023lost} & Low self-esteem & Reddit posts with classification labels & P:~2{,}926 & P:~162 & Source records \\
Twitter ST-MHD \citep{suhavi2022twitterstmhd} & Mental-health conditions & Tweets with condition labels & P:~2{,}013 & P:~108 & Authors \\
\midrule
\multicolumn{6}{@{}l}{\textbf{Questionnaire and survey records}}\\*
\midrule
\noalign{\nobreak}
16PF \citep{openpsychometrics} & Sixteen personality factors & Questionnaire responses & M:~1{,}727\newline P:~12{,}000 & P:~200 & Respondents \\
AS+\allowbreak SC+\allowbreak AD+\allowbreak DO \citep{openpsychometrics} & Assertiveness, social confidence, adventurousness and dominance & Questionnaire responses & M:~1{,}781\newline P:~1{,}800 & P:~50 & Respondents \\
Big Five \citep{goldberg2006ipip,openpsychometrics} & Five-factor personality & Questionnaire responses & M:~2{,}000\newline P:~12{,}000 & P:~200 & Respondents \\
CFCS \citep{openpsychometrics} & Consideration of future consequences & Questionnaire responses & P:~12{,}000 & P:~200 & Respondents \\
DASS \citep{lovibond1995dass,openpsychometrics} & Depression, anxiety and stress & Questionnaire responses & M:~2{,}000\newline P:~12{,}000 & P:~200 & Respondents \\
ECR \citep{brennan1998attachment,openpsychometrics} & Attachment anxiety and avoidance & Questionnaire responses & M:~1{,}972\newline P:~12{,}000 & P:~200 & Respondents \\
FTI \citep{openpsychometrics} & Temperament styles & Questionnaire responses & P:~4{,}471 & P:~200 & Respondents \\
HBDS \citep{openpsychometrics} & Self-reported cognitive preferences & Questionnaire responses & P:~12{,}000 & P:~200 & Respondents \\
HSNS+DD \citep{openpsychometrics} & Narcissism and antagonistic traits & Questionnaire responses & M:~1{,}945\newline P:~12{,}000 & P:~200 & Respondents \\
HSQ \citep{openpsychometrics} & Affiliative, self-enhancing, aggressive and self-defeating humour & Questionnaire responses & P:~1{,}924 & P:~54 & Respondents \\
IPIP-FFM \citep{goldberg2006ipip,openpsychometrics} & Five-factor personality & Questionnaire responses & M:~2{,}000\newline P:~12{,}000 & P:~200 & Respondents \\
NPAS \citep{openpsychometrics} & Intellectual interests and preferences & Questionnaire responses & P:~12{,}000 & P:~200 & Respondents \\
NPI \citep{openpsychometrics} & Narcissistic traits & Questionnaire responses & P:~10{,}112 & P:~200 & Respondents \\
NR-6 \citep{openpsychometrics} & Connection to nature & Questionnaire responses & P:~2{,}006 & P:~111 & Respondents \\
OSRI-44 \citep{openpsychometrics} & Gender-role traits & Questionnaire responses & P:~12{,}000 & P:~200 & Respondents \\
PWE \citep{openpsychometrics} & Work-related beliefs and values & Questionnaire responses & P:~1{,}215 & P:~67 & Respondents \\
RSE \citep{rosenberg1965selfesteem,openpsychometrics} & Global self-esteem & Questionnaire responses & M:~2{,}000\newline P:~12{,}000 & P:~200 & Respondents \\
SD3 \citep{jones2014sd3,openpsychometrics} & Machiavellianism, narcissism and psychopathy & Questionnaire responses & M:~2{,}000\newline P:~12{,}000 & P:~200 & Respondents \\
World Values Survey \citep{haerpfer2022wvs7} & Social, political, moral and religious attitudes & Survey responses & P:~12{,}000 & P:~200 & Respondents \\
HEXACO \citep{lee2004hexaco,openpsychometrics} & Six-factor personality & Questionnaire responses & M:~2{,}000 & P:~200 & Domain excluded from P training \\
MACH-IV \citep{openpsychometrics} & Manipulative and cynical interpersonal attitudes & Questionnaire responses & M:~1{,}999 & P:~200 & Domain excluded from P training \\
RIASEC \citep{holland1997vocational,openpsychometrics} & Vocational interests & Questionnaire responses & M:~2{,}000 & P:~200 & Domain excluded from P training \\
SAPA-2022 \citep{condon2024sapa} & Personality traits and facets & Questionnaire responses & --- & P:~200 & Domain excluded from P training \\
European Social Survey Round 9 (ESS09) \citep{ess2018round9} & All 109 constructs (nomological validity) & Survey responses & --- & M:~4{,}921 & Evaluation only \\
\midrule
\multicolumn{6}{@{}l}{\textbf{Synthetic measurement records}}\\*
\midrule
\noalign{\nobreak}
Synthetic questionnaire records & All 109 constructs & Simulated item responses with instrument-scored targets & M:~10{,}900 & --- & Training only \\
Calibrated synthetic narratives & All 109 constructs & First-person narratives with comparison-derived scores & M:~15{,}250 & --- & Training only \\
\midrule
\multicolumn{6}{@{}l}{\textbf{Social science experiments}}\\*
\midrule
\noalign{\nobreak}
Management studies (AMJ) \citep{cui2025largescale} & Management and organizational behaviour & Experimental scenarios, human statistics and archived model responses & Pop:~3{,}790\newline (9 studies) & Pop:~896\newline (4 studies) & Studies \\
Applied psychology studies (JAP) \citep{cui2025largescale} & Applied and organizational psychology & Experimental scenarios, human statistics and archived model responses & Pop:~2{,}552\newline (7 studies) & Pop:~755\newline (2 studies) & Studies \\
Experimental psychology studies (JEP) \citep{cui2025largescale} & Experimental psychological findings & Experimental scenarios, human statistics and archived model responses & Pop:~4{,}097\newline (16 studies) & Pop:~499\newline (1 study) & Studies \\
Social and personality psychology studies (JPSP) \citep{cui2025largescale} & Social behaviour and personality & Experimental scenarios, human statistics and archived model responses & Pop:~6{,}970\newline (21 studies) & Pop:~955\newline (4 studies) & Studies \\
Organizational behaviour and decision studies (OBHDP) \citep{cui2025largescale} & Organizational behaviour and human decision processes & Experimental scenarios, human statistics and archived model responses & Pop:~5{,}782\newline (10 studies) & Pop:~1{,}172\newline (4 studies) & Studies \\
\midrule
\multicolumn{3}{@{}l}{\textbf{Total by objective}} & \textbf{M:~53{,}073}\newline \textbf{P:~192{,}179}\newline \textbf{Pop:~23{,}191} & \textbf{M:~10{,}820}\newline \textbf{P:~4{,}767}\newline \textbf{Pop:~4{,}277} & \\
\end{longtable}
}

\subsection{Social science experiments}\label{app:data-population}

We use the dataset released by \citet{cui2025largescale}, which provides experimental materials, published human summary statistics and archived LLM responses for scenario-based studies in psychology and management. We select studies whose experimental materials and response tasks can be presented entirely in text and completed without a multi-round reasoning or interaction procedure. The selected studies span management, applied psychology, experimental psychology, social and personality psychology, and organizational behaviour, and decision-making. \Cref{tab:data-population} lists the studies, split assignments, and data quantities, where study identifiers follow \citet{cui2025largescale}.

\begin{longtable}{@{}llrrrr@{}}
\caption{Social science experiments used for population alignment and simulation. We report the split assignment, number of experimental conditions, virtual respondents (synthesized for training or simulated per model for evaluation), individual item responses (decisions), and experimental effects.}\label{tab:data-population}\\
\toprule
Study & Split & Conditions & Respondents & Decisions & Effects \\
\midrule\endfirsthead
\multicolumn{6}{@{}l}{\emph{Table \ref{tab:data-population} continued}}\\
\toprule
Study & Split & Conditions & Respondents & Decisions & Effects \\
\midrule\endhead
\midrule\multicolumn{6}{r}{\emph{continued on next page}}\\\endfoot
\bottomrule\endlastfoot
\multicolumn{6}{@{}l}{\textbf{Train studies} (63)}\\
\midrule
\texttt{AMJ\_10\_2} & train & 4 & 196 & 2,940 & 3 \\
\texttt{AMJ\_1\_3} & train & 10 & 922 & 5,532 & 2 \\
\texttt{AMJ\_2\_2} & train & 6 & 307 & 3,070 & 3 \\
\texttt{AMJ\_3\_2} & train & 4 & 78 & 78 & 2 \\
\texttt{AMJ\_3\_3} & train & 4 & 122 & 2,806 & 7 \\
\texttt{AMJ\_4\_3} & train & 4 & 440 & 7,480 & 4 \\
\texttt{AMJ\_5\_1} & train & 6 & 1,189 & 32,103 & 11 \\
\texttt{AMJ\_8\_2} & train & 3 & 336 & 6,720 & 4 \\
\texttt{AMJ\_9\_1} & train & 4 & 200 & 7,200 & 2 \\
\texttt{JAP\_3\_1c} & train & 4 & 169 & 1,521 & 2 \\
\texttt{JAP\_5\_1} & train & 8 & 205 & 3,690 & 5 \\
\texttt{JAP\_6\_3} & train & 6 & 671 & 10,065 & 8 \\
\texttt{JAP\_6\_4} & train & 6 & 1,104 & 9,936 & 25 \\
\texttt{JAP\_7\_3} & train & 4 & 134 & 1,072 & 6 \\
\texttt{JAP\_8\_1} & train & 6 & 129 & 1,548 & 6 \\
\texttt{JAP\_8\_2} & train & 6 & 140 & 2,100 & 6 \\
\texttt{JEP\_10\_2} & train & 3 & 249 & 3,984 & 8 \\
\texttt{JEP\_1\_1a} & train & 6 & 301 & 301 & 3 \\
\texttt{JEP\_1\_1b} & train & 4 & 430 & 430 & 3 \\
\texttt{JEP\_1\_3} & train & 4 & 401 & 2,005 & 7 \\
\texttt{JEP\_1\_6b} & train & 6 & 525 & 2,625 & 12 \\
\texttt{JEP\_2\_1} & train & 2 & 363 & 4,356 & 6 \\
\texttt{JEP\_2\_2a} & train & 2 & 104 & 832 & 1 \\
\texttt{JEP\_2\_2b} & train & 2 & 118 & 944 & 1 \\
\texttt{JEP\_3\_1} & train & 3 & 204 & 816 & 3 \\
\texttt{JEP\_4\_2} & train & 4 & 127 & 127 & 3 \\
\texttt{JEP\_4\_3} & train & 4 & 96 & 96 & 3 \\
\texttt{JEP\_5\_1b} & train & 2 & 154 & 462 & 2 \\
\texttt{JEP\_5\_3a} & train & 2 & 142 & 852 & 1 \\
\texttt{JEP\_5\_3b} & train & 2 & 260 & 1,820 & 6 \\
\texttt{JEP\_8\_4} & train & 2 & 194 & 1,358 & 5 \\
\texttt{JEP\_9\_1} & train & 2 & 429 & 4,719 & 4 \\
\texttt{JPSP\_10\_1} & train & 4 & 205 & 3,895 & 3 \\
\texttt{JPSP\_10\_2} & train & 4 & 200 & 4,000 & 3 \\
\texttt{JPSP\_10\_3} & train & 4 & 197 & 2,955 & 3 \\
\texttt{JPSP\_10\_4} & train & 4 & 198 & 4,752 & 3 \\
\texttt{JPSP\_1\_1} & train & 2 & 310 & 5,580 & 2 \\
\texttt{JPSP\_1\_2} & train & 4 & 156 & 2,964 & 2 \\
\texttt{JPSP\_1\_4} & train & 4 & 409 & 9,816 & 3 \\
\texttt{JPSP\_1\_5} & train & 3 & 1,008 & 30,240 & 11 \\
\texttt{JPSP\_3\_4} & train & 6 & 346 & 7,612 & 10 \\
\texttt{JPSP\_4\_5} & train & 3 & 584 & 7,008 & 10 \\
\texttt{JPSP\_4\_7} & train & 2 & 418 & 2,090 & 1 \\
\texttt{JPSP\_6\_1a} & train & 2 & 144 & 432 & 1 \\
\texttt{JPSP\_6\_1b} & train & 2 & 267 & 801 & 1 \\
\texttt{JPSP\_6\_1c} & train & 3 & 209 & 1,045 & 12 \\
\texttt{JPSP\_6\_3} & train & 4 & 353 & 2,118 & 3 \\
\texttt{JPSP\_6\_4a} & train & 4 & 352 & 1,056 & 3 \\
\texttt{JPSP\_6\_4b} & train & 6 & 532 & 1,596 & 4 \\
\texttt{JPSP\_6\_5} & train & 4 & 343 & 1,029 & 3 \\
\texttt{JPSP\_8\_1} & train & 2 & 209 & 1,672 & 1 \\
\texttt{JPSP\_8\_2} & train & 4 & 410 & 3,280 & 6 \\
\texttt{JPSP\_8\_4} & train & 2 & 120 & 840 & 1 \\
\texttt{OBHDP\_3\_3} & train & 8 & 219 & 438 & 13 \\
\texttt{OBHDP\_3\_6} & train & 4 & 1,599 & 9,594 & 16 \\
\texttt{OBHDP\_5\_2} & train & 2 & 442 & 3,536 & 3 \\
\texttt{OBHDP\_6\_2} & train & 4 & 1,654 & 19,848 & 4 \\
\texttt{OBHDP\_6\_3} & train & 4 & 189 & 1,890 & 4 \\
\texttt{OBHDP\_6\_4} & train & 2 & 197 & 1,970 & 3 \\
\texttt{OBHDP\_7\_1} & train & 4 & 438 & 1,314 & 6 \\
\texttt{OBHDP\_7\_2b} & train & 3 & 405 & 7,695 & 14 \\
\texttt{OBHDP\_9\_2} & train & 4 & 310 & 1,860 & 3 \\
\texttt{OBHDP\_9\_3} & train & 4 & 329 & 2,961 & 3 \\
\midrule
\multicolumn{2}{@{}l}{\emph{Subtotal}} & 247 & 23,191 & 269,475 & 319 \\
\midrule
\multicolumn{6}{@{}l}{\textbf{Held-out test studies} (15)}\\
\midrule
\texttt{AMJ\_4\_2} & test & 2 & 163 & 2,282 & 2 \\
\texttt{AMJ\_6\_2} & test & 4 & 299 & 4,485 & 4 \\
\texttt{AMJ\_6\_3} & test & 4 & 309 & 4,635 & 4 \\
\texttt{AMJ\_7\_1} & test & 4 & 125 & 1,000 & 3 \\
\texttt{JAP\_8\_4} & test & 4 & 550 & 6,600 & 6 \\
\texttt{JAP\_9\_4} & test & 4 & 205 & 2,870 & 7 \\
\texttt{JEP\_1\_6a} & test & 4 & 499 & 1,996 & 2 \\
\texttt{JPSP\_4\_2} & test & 2 & 133 & 665 & 1 \\
\texttt{JPSP\_4\_3} & test & 2 & 209 & 2,508 & 3 \\
\texttt{JPSP\_4\_4} & test & 2 & 204 & 816 & 2 \\
\texttt{JPSP\_4\_8} & test & 2 & 409 & 2,863 & 3 \\
\texttt{OBHDP\_10\_1b} & test & 2 & 202 & 606 & 1 \\
\texttt{OBHDP\_10\_1c} & test & 4 & 405 & 1,215 & 3 \\
\texttt{OBHDP\_10\_4} & test & 2 & 199 & 398 & 2 \\
\texttt{OBHDP\_7\_2a} & test & 3 & 366 & 5,490 & 10 \\
\midrule
\multicolumn{2}{@{}l}{\emph{Subtotal}} & 45 & 4,277 & 38,429 & 53 \\
\end{longtable}

\subsection{Psychological constructs}\label{app:constructs}
\Cref{tab:constructs} lists the 109 constructs used to train the measurement head, with their definitions and references. Each definition describes the intended meaning of a high construct score. The reference column identifies the conceptual basis, not necessarily the item instrument used in training or evaluation.

\paragraph{Personality and vocational interests.}
IPIP and IPIP\_NEO refer to the International Personality Item Pool, a public-domain item repository spanning personality, interpersonal, affective and cognitive scales \citep{goldberg2006ipip,ipipScaleIndex}; HEXACO to the HEXACO Personality Inventory \citep{lee2004hexaco,hexacoScaleDescriptions}; SD3 to the Short Dark Triad \citep{jones2014sd3}; and RIASEC to Holland's vocational interest model \citep{holland1997vocational}.

\paragraph{Self-evaluation and interpersonal attributes.}
RSES refers to the Rosenberg Self-Esteem Scale \citep{rosenberg1965selfesteem}; ECR to the Experiences in Close Relationships scale \citep{brennan1998attachment}; and IRI to the Interpersonal Reactivity Index \citep{davis1983iri}, whose Empathic Concern and Perspective Taking subscales capture affective and cognitive empathy. IRI also provides the conceptual reference for the separate general Empathy construct.

\paragraph{Affect, coping and wellbeing.}
DASS refers to the Depression Anxiety Stress Scales \citep{lovibond1995dass}; EAC to the Emotional Approach Coping scale \citep{stanton2000eac}; ERQ to the Emotion Regulation Questionnaire \citep{gross2003erq}; RESILIENCE to the Connor--Davidson Resilience Scale \citep{connor2003resilience}; SWLS to the Satisfaction With Life Scale \citep{diener1985swls}; SCSQ to the Simplified Coping Style Questionnaire \citep{xie1998scsq}; WORKAHOLISM to the Dutch Work Addiction Scale \citep{schaufeli2009workaholism}; and ASRS to the Adult ADHD Self-Report Scale \citep{kessler2005asrs}, used here as a definition reference for \texttt{adhd}.

\paragraph{Values and character strengths.}
SCHWARTZ refers to Schwartz's theory of basic human values \citep{schwartz2012values}; MFQ2 to the Moral Foundations Questionnaire-2 \citep{atari2023mfq2}; HOFSTEDE to Hofstede's Values Survey Module \citep{hofstede2013vsm} (used here to score individual value orientations rather than national cultures); LVI to the Life Values Inventory \citep{brown1996lvi}; and VIA and VIA\_IPIP to the Values in Action classification of character strengths and its IPIP scales \citep{peterson2004strengths,ipipScaleIndex}.

\paragraph{Other dispositional constructs.}
CATPD refers to the Computerized Adaptive Test of Personality Disorder \citep{simms2011catpd}; MPQ to the Multidimensional Personality Questionnaire \citep{mpq2023}; BISBAS to the Behavioral Inhibition and Behavioral Activation Scales \citep{carver1994bisbas}; NFC to the Need for Cognition scale \citep{cacioppo1982nfc}; and SELF\_EFFICACY and LOCUS to Bandura's self-efficacy theory and Rotter's locus of control scale \citep{bandura1977selfefficacy,rotter1966locus}.

{
\begin{longtable}{@{}rllp{0.46\columnwidth}@{}}
\caption{The 109 supervised psychological constructs.}\label{tab:constructs}\\
\toprule
\# & Construct & Reference & Definition \\
\midrule
\endfirsthead
\multicolumn{4}{@{}l}{\emph{Table \ref{tab:constructs} continued}}\\
\toprule
\# & Construct & Reference & Definition \\
\midrule
\endhead
\midrule\multicolumn{4}{r}{\emph{continued on next page}}\\
\endfoot
\bottomrule
\endlastfoot
\multicolumn{4}{@{}l}{\textbf{Broad personality and vocational interests} (24)}\\
\midrule
1 & \texttt{openness} & IPIP & A stable tendency to be curious about and receptive to new ideas, aesthetics, imagination, and varied experiences. \\
2 & \texttt{conscientiousness} & IPIP & A stable tendency to be organized, self-disciplined, dependable, and persistent in pursuing goals and standards. \\
3 & \texttt{extraversion} & IPIP & A tendency to seek and enjoy social interaction and high activity, expressed through energy, enthusiasm, and sociability. \\
4 & \texttt{agreeableness} & IPIP & A stable tendency to treat others with trust, cooperation, consideration, and tolerance. \\
5 & \texttt{neuroticism} & IPIP & A tendency to experience negative emotions and stress reactions, such as anxiety, sadness, and anger, frequently and intensely. \\
6 & \texttt{machiavellianism} & SD3 & A continuous tendency to pursue self-interest through strategic manipulation, cynicism, and instrumental treatment of others. \\
7 & \texttt{narcissism} & SD3 & A nonclinical continuous tendency toward inflated self-importance, admiration seeking, superiority, and entitlement. \\
8 & \texttt{psychopathy} & SD3 & A nonclinical continuous tendency toward callousness, low empathy, manipulation, impulsive risk taking, and disregard for norms. \\
9 & \texttt{realistic} & RIASEC & An interest in hands-on, mechanical, outdoor, or practical activities that produce tangible results. \\
10 & \texttt{investigative} & RIASEC & An interest in observing, analyzing, researching, and solving scientific or complex problems. \\
11 & \texttt{artistic} & RIASEC & An interest in creativity, aesthetics, self-expression, and activities with relatively little imposed structure. \\
12 & \texttt{enterprising} & RIASEC & An interest in persuading, leading, selling, entrepreneuring, or organizing others to achieve goals. \\
13 & \texttt{conventional} & RIASEC & An interest in rule-governed, detail-oriented activities involving data, records, and structured procedures. \\
14 & \texttt{assertiveness} & IPIP & A tendency to state opinions and needs openly, speak up, and take initiative in interactions. \\
15 & \texttt{dominance} & IPIP & A tendency to exert influence, control decisions, and seek status in interpersonal or group settings. \\
16 & \texttt{honesty humility} & HEXACO & A tendency toward sincerity, fairness, and modesty, with little willingness to exploit others through deceit, greed, or status privilege. \\
17 & \texttt{emotionality} & HEXACO & A tendency toward fear and worry, need for emotional support, and strong sentimental attachment to close others. \\
18 & \texttt{warmth} & IPIP & A tendency to show affection, concern, emotional engagement, and interest in close connections with others. \\
19 & \texttt{emotional stability} & IPIP & A tendency to remain composed, recover equilibrium, and show relatively little emotional volatility under stress or setbacks. \\
20 & \texttt{rule consciousness} & IPIP & A tendency to value and conscientiously follow rules, duties, and social standards. \\
21 & \texttt{self reliance} & IPIP & A preference for independent decisions and action, with comfort acting alone and less reliance on group support. \\
22 & \texttt{perfectionism} & IPIP & A tendency to demand precision, order, high standards, and error-free outcomes. \\
23 & \texttt{Risk Taking} & CATPD & A willingness to choose actions with uncertain outcomes and possible loss or danger in pursuit of reward. \\
24 & \texttt{humor playfulness} & VIA\_IPIP & A tendency to joke, amuse others, play, and enjoy life lightly while creating a lively and pleasant atmosphere. \\
\midrule
\multicolumn{4}{@{}l}{\textbf{Self-evaluation, interpersonal attachment, and social functioning} (22)}\\
\midrule
25 & \texttt{self esteem} & RSES & A person's overall positive evaluation of and respect for their own worth. \\
26 & \texttt{avoidance} & ECR & In attachment relationships, discomfort with intimacy and dependence and a preference for emotional distance and excessive self-reliance. \\
27 & \texttt{Empathic Concern} & IRI & A tendency to experience warmth, compassion, and protective concern when other people suffer, face misfortune, or are treated unfairly. \\
28 & \texttt{Perspective Taking} & IRI & A tendency to adopt other people's psychological viewpoints spontaneously in everyday interactions. \\
29 & \texttt{Forgiveness} & IPIP & A tendency to reduce resentment and retaliatory motivation after an offense and gradually restore goodwill. \\
30 & \texttt{Social Self-Esteem} & HEXACO & A positive evaluation of being liked, accepted, and effective in social interaction. \\
31 & \texttt{Sociability} & IPIP & A tendency to prefer, enjoy, and actively seek the company of other people. \\
32 & \texttt{Altruism} & IPIP & A willingness to help others without direct reward and sometimes at personal cost. \\
33 & \texttt{Self-disclosure} & IPIP & The degree of willingness to reveal personal experiences, thoughts, and feelings to others. \\
34 & \texttt{Leadership} & IPIP & A tendency to accept responsibility, coordinate groups, and influence others toward shared goals. \\
35 & \texttt{Empathy} & IRI & The capacity and tendency to understand another person's emotions and situation and respond with corresponding concern or feeling. \\
36 & \texttt{Cooperation} & IPIP & A tendency to coordinate, compromise, share responsibility, and work jointly with others toward common goals. \\
37 & \texttt{Callousness} & CATPD & A lack of sympathy, guilt, or concern regarding other people's suffering and needs. \\
38 & \texttt{Manipulativeness} & CATPD & A tendency to use deception, charm, or pressure to control others for personal benefit. \\
39 & \texttt{Submissiveness} & CATPD & A tendency to yield one's decisions and needs and to defer or become controlled in interpersonal relationships. \\
40 & \texttt{Social Withdrawal} & CATPD & A tendency to avoid or disengage from social contact, close relationships, and group participation. \\
41 & \texttt{Social Closeness} & MPQ & A tendency to value and enjoy close interpersonal bonds, prefer being with others, and turn to others for comfort when needed. \\
42 & \texttt{Aggression} & IPIP & A tendency to respond to conflict or frustration with hostile, threatening, or harmful verbal or physical behavior. \\
43 & \texttt{Belonging} & LVI & The degree to which being accepted, liked, and included by others is treated as an important life value. \\
44 & \texttt{trust} & IPIP & A tendency to expect that other people are generally honest, dependable, and unlikely to cause deliberate harm. \\
45 & \texttt{gratitude} & VIA & A tendency to notice and appreciate benefits received and to feel thankful toward benefactors or life circumstances. \\
46 & \texttt{teamwork citizenship} & VIA & A tendency to identify with a group, fulfill responsibilities reliably, and contribute to the common good. \\
\midrule
\multicolumn{4}{@{}l}{\textbf{Affect, mental health, and wellbeing} (22)}\\
\midrule
47 & \texttt{depression} & DASS & A continuous state characterized by dysphoria, hopelessness, anhedonia, low self-worth, and low motivation. \\
48 & \texttt{anxiety} & DASS & A continuous state characterized by physiological arousal, fear, situational anxiety, and apprehension about threat. \\
49 & \texttt{stress} & DASS & A continuous state of tension, difficulty relaxing, irritability, and heightened reactivity. \\
50 & \texttt{Emotional Processing} & EAC & A coping tendency to acknowledge, explore, and understand one's emotions and their meaning actively. \\
51 & \texttt{Fearfulness} & HEXACO & A tendency to experience fear and avoidance in response to physical danger, injury, or threat. \\
52 & \texttt{Cognitive reappraisal} & ERQ & An emotion-regulation strategy that changes a situation's interpretation in order to alter its emotional impact. \\
53 & \texttt{Expressive suppression} & ERQ & An emotion-regulation strategy that inhibits outward facial, vocal, or behavioral expression after emotion has arisen. \\
54 & \texttt{Resilience} & RESILIENCE & The capacity to maintain adaptation and recover functioning when facing adversity, stress, or change. \\
55 & \texttt{Satisfaction with life} & SWLS & A cognitive judgment of how well one's life as a whole meets one's personal standards. \\
56 & \texttt{Positive coping} & SCSQ & The use of adaptive responses to stress, including problem solving, seeking support, and positive reappraisal. \\
57 & \texttt{Workaholism} & WORKAHOLISM & Working excessively beyond external requirements while feeling internally compelled and unable to disengage. \\
58 & \texttt{Emotional Detachment} & CATPD & A tendency to restrict emotional involvement and expression and remain distant from intimacy or emotional connection. \\
59 & \texttt{Cognitive Problems} & CATPD & Recurrent difficulties with attention, memory, continuity of thought, or organization of thinking. \\
60 & \texttt{Affective Lability} & CATPD & A tendency for emotional states to shift rapidly, intensely, and unpredictably. \\
61 & \texttt{Anger} & IPIP & A tendency to experience irritation and hostile affect frequently or intensely in response to frustration, offense, or unfairness. \\
62 & \texttt{Wellbeing} & MPQ & A cheerful, enthusiastic, and optimistic disposition, accompanied by feeling good about oneself and expecting a positive future. \\
63 & \texttt{vitality enthusiasm zest} & VIA & A tendency to approach daily activities and life with energy, liveliness, and enthusiasm. \\
64 & \texttt{hope optimism} & IPIP & Positive expectations about the future together with confidence that routes toward goals can be found. \\
65 & \texttt{rumination} & IPIP & Repetitive, passive focus on distress and its causes or consequences rather than effective problem resolution. \\
66 & \texttt{self acceptance} & IPIP & A realistic acknowledgment of one's strengths, limitations, and past accompanied by an overall positive attitude toward oneself. \\
67 & \texttt{self consciousness} & IPIP\_NEO & A tendency to feel embarrassed, intimidated, and uncomfortable under social attention or evaluation and to worry about behaving inadequately. \\
68 & \texttt{adhd} & ASRS & A continuous tendency toward adult inattentive and/or hyperactive-impulsive symptoms; it does not by itself indicate a diagnosis. \\
\midrule
\multicolumn{4}{@{}l}{\textbf{Values, morality, culture, and worldviews} (23)}\\
\midrule
69 & \texttt{Tradition} & SCHWARTZ & Valuing respect for, acceptance of, and commitment to customs and ideas inherited from traditional culture or religion. \\
70 & \texttt{Individualism} & HOFSTEDE & As a cultural value orientation, prioritizing an independent self, personal choice, and individual goals over strong group embeddedness. \\
71 & \texttt{Power Distance} & HOFSTEDE & As a cultural value orientation, accepting unequal distributions of power and status in organizations or society. \\
72 & \texttt{Indulgence} & HOFSTEDE & As a cultural value orientation, allowing relatively free gratification of desires related to leisure, enjoyment, and having fun rather than strict restraint. \\
73 & \texttt{Long Term Orientation} & HOFSTEDE & As a cultural value orientation, fostering adaptation, perseverance, and thrift toward future rewards rather than primarily emphasizing tradition, face, and present social obligations. \\
74 & \texttt{Uncertainty Avoidance} & HOFSTEDE & As a cultural value orientation, feeling threatened by ambiguity and the unknown and relying on rules and predictability to reduce uncertainty. \\
75 & \texttt{Self-Direction} & SCHWARTZ & Valuing independent thought, free choice, creativity, and exploration. \\
76 & \texttt{Power} & SCHWARTZ & Valuing social status, prestige, and control over people or resources. \\
77 & \texttt{Universalism} & SCHWARTZ & Valuing understanding, tolerance, and protection of the welfare of all people and of nature. \\
78 & \texttt{Achievement} & SCHWARTZ & Valuing personal success through demonstrating competence according to socially recognized standards. \\
79 & \texttt{Security} & SCHWARTZ & Valuing safety, harmony, and stability for oneself, one's relationships, and society. \\
80 & \texttt{Stimulation} & SCHWARTZ & Valuing excitement, novelty, and challenge in life. \\
81 & \texttt{Conformity} & SCHWARTZ & Restraining actions likely to harm others, disrupt order, or violate social expectations and norms. \\
82 & \texttt{Hedonism} & SCHWARTZ & Valuing pleasure, enjoyment, and sensuous gratification. \\
83 & \texttt{Benevolence} & SCHWARTZ & Valuing the preservation and enhancement of the welfare of close others with whom one has frequent contact. \\
84 & \texttt{Loyalty} & MFQ2 & Treating solidarity, allegiance, and self-sacrifice for one's group, community, or nation as moral obligations. \\
85 & \texttt{Authority} & MFQ2 & Treating respect for legitimate hierarchy and leadership and maintenance of social order as moral obligations. \\
86 & \texttt{Purity} & MFQ2 & Treating sanctity, self-control, and avoidance of physical or spiritual degradation as moral concerns. \\
87 & \texttt{Equality} & MFQ2 & Treating equal rights, treatment, and outcomes as important bases of fairness. \\
88 & \texttt{Proportionality} & MFQ2 & Treating rewards and punishments proportional to contribution, merit, or responsibility as a basis of fairness. \\
89 & \texttt{Morality} & IPIP & A tendency to follow basic ethical principles, act honestly and fairly, and avoid benefiting through deception or harm to others. \\
90 & \texttt{Spirituality} & LVI & Valuing sacred or transcendent meaning, beliefs, and connection to something larger than the everyday self. \\
91 & \texttt{Masculinity} & HOFSTEDE & As a cultural value orientation, emphasizing assertiveness, toughness, achievement, and material success over modesty, care, and quality of life. \\
\midrule
\multicolumn{4}{@{}l}{\textbf{Cognition, self-regulation, and agency} (18)}\\
\midrule
92 & \texttt{Planning} & IPIP & A tendency to set goals, organize steps and resources in advance, and consider the future consequences of action. \\
93 & \texttt{Patience} & HEXACO & A tendency to remain calm during delays, obstacles, or provocation rather than becoming impatient or angry. \\
94 & \texttt{Industriousness} & IPIP & A tendency to invest sustained effort and persist in completing difficult or time-consuming work. \\
95 & \texttt{Autonomy} & IPIP & A tendency to act from one's own judgment and resist inappropriate social pressure or excessive dependence. \\
96 & \texttt{Impulse-Control} & IPIP & The capacity to inhibit immediate urges and unconsidered reactions in service of longer-term goals or standards. \\
97 & \texttt{Imagination} & IPIP & A tendency to form vivid and novel mental representations and envision possibilities beyond present reality. \\
98 & \texttt{Competence} & IPIP & The perceived and demonstrated capacity to understand tasks, apply knowledge, and meet challenges effectively. \\
99 & \texttt{Drive} & BISBAS & A tendency to pursue desired rewards or goals actively and persistently. \\
100 & \texttt{Reward Responsiveness} & BISBAS & A tendency to experience positive affect and approach motivation when rewards are anticipated or obtained. \\
101 & \texttt{Harm Avoidance} & IPIP & A tendency toward caution, avoidance, and behavioral inhibition when danger or harm is anticipated. \\
102 & \texttt{intellect} & IPIP & A trait-like enjoyment of abstract reasoning and complex ideas; it is not equivalent to measured intelligence. \\
103 & \texttt{creativity originality} & IPIP & A tendency to produce ideas, expressions, or solutions that are both novel and appropriate to the context. \\
104 & \texttt{need for cognition} & NFC & A stable tendency to engage in and enjoy effortful thinking. \\
105 & \texttt{self discipline} & IPIP & A tendency to initiate and persist in goal-relevant tasks despite boredom, frustration, or distraction. \\
106 & \texttt{self efficacy} & SELF\_EFFICACY & A belief in one's capability to organize and execute the actions required to handle particular situations. \\
107 & \texttt{curiosity} & VIA\_IPIP & A tendency to explore new information actively, ask questions, seek explanations, and enjoy discovering and understanding new things. \\
108 & \texttt{problem solving} & IPIP & A tendency to identify a problem's structure, analyze causes, compare alternatives, and develop workable solutions. \\
109 & \texttt{locus of control} & LOCUS & The extent to which outcomes are expected to depend on one's own actions rather than luck, fate, or powerful others. \\
\end{longtable}
}%

\FloatBarrier

\section{Implementation details}\label{app:implementation}

\subsection{Data synthesis for measurement training}\label{app:data-construction}
\paragraph{Synthetic questionnaires.}
We construct 10{,}900 synthetic questionnaire records to train the measurement head. For each record, we sample up to ten distinct items from one construct's item pool and then generate their responses. To keep responses within a record broadly consistent, we first draw a value from a uniform distribution to represent the record's intended construct level. We then use Gaussian-based sampling around this shared value to select a response to each item, accounting for its scoring direction and available response options. This introduces variation across individual answers while preserving a common underlying tendency. In 10\% of records, responses are instead sampled independently and uniformly from the available options. Once the responses have been generated, we calculate the supervision target by mapping them to $[-1,1]$, applying reverse scoring where required, and averaging the item scores. The resulting records are presented either as explicit item--response pairs or as natural-language statements produced by fixed templates. Finally, we remove duplicate records and sample across score bins to retain 100 records per construct, covering the full range of construct levels.

\paragraph{Synthetic narratives.}
We use DeepSeek-V4-Pro to generate 25{,}000 candidate narratives across the 109 constructs. Each request supplies a construct definition together with generation conditions grounded in real-world scenarios. We balance the six combinations of aligned or opposing direction and weak, medium or strong evidence. The texts cover ten everyday contexts, including work or study, family life, social interactions, online communication, leisure, daily routines, conflict, setbacks and personal decisions. They take the form of concrete episodes, preferences or opinions, and self-reflections in a 5:3:2 ratio. Requested lengths are 8--99, 100--250 and 250--500 words, in a 3:3:2 ratio. The model writes each narrative independently and is instructed to express the relevant behaviour or experience without explicitly naming the psychological construct or reporting a score. The generation system and user prompts are given below.

\begin{promptbox}[title={System prompt}]
You write natural English posts and comments for a controlled text dataset.\par
\medskip
INPUT FORMAT\par
The user supplies one target construct, its aligned-direction definition, and a JSON\par
array of slots. Each slot contains:\par
- slot\_id: an identifier that must be copied exactly;\par
- target\_direction: "aligned" or "opposing";\par
- strength: "weak", "medium", or "strong";\par
- context: the setting for the text;\par
- expression\_mode: the requested form of writing;\par
- word\_range: the requested inclusive word-count range.\par
\medskip
OUTPUT FORMAT\par
Return only one valid JSON object with exactly this shape:\par
\{"items": [\{"slot\_id": "...", "text": "..."\}]\}\par
\medskip
Return exactly one item for every input slot, in the same order, and no extra\par
items. Each item must contain only slot\_id and text. Preserve every slot\_id\par
exactly. The text must be a JSON string with no surrounding commentary.\par
\medskip
CONTENT RULES\par
The text is the author's post itself, not a description of a post. Never include\par
a psychological label, score, intensity label, explanation, heading, bullet, or\par
wrapper such as "I posted on Reddit". Ordinary words that happen to overlap a\par
construct name may still be used naturally; do not explicitly classify the\par
author. Do not imitate a questionnaire item.\par
\medskip
Use plausible everyday language rather than polished fiction. Every text must\par
fall within its requested inclusive word\_range; count words as\par
whitespace-separated tokens. Before returning, verify the length and expand or\par
trim neutral setting and narrative detail as needed without changing the\par
requested direction or evidential strength. Let sentence count and structure\par
follow the requested word\_range and the natural content.\par
\medskip
Each slot has four independent controls:\par
- target\_direction: whether the text should indicate the "aligned" direction\par
  described by the definition or its "opposing" direction;\par
- strength: how diagnostic the evidence should be;\par
- context: the ordinary setting in which the evidence appears;\par
- expression\_mode: the linguistic form in which the evidence is expressed.\par
\medskip
Strength describes diagnostic evidence, not dramatic wording:\par
- weak: exactly one mild, situational, directionally interpretable clue; do not\par
  reinforce it with repeated examples, broad self-descriptions, strong emotion,\par
  or a consequential choice;\par
- medium: clear ordinary evidence, stronger than a passing clue but limited in\par
  scope and without exaggeration;\par
- strong: decisive, repeated, pervasive, or consequential evidence that remains\par
  realistic.\par
\medskip
Apply these distinctions symmetrically to both directions. For each strength, use\par
only the corresponding level of diagnostic evidence described above in the\par
requested direction. Derive all construct-specific content, including opposing\par
implications, solely from the supplied definition.\par
\medskip
Word count controls narrative length, not evidential strength. Longer texts may\par
add setting and neutral detail, but must not accumulate extra construct-relevant\par
examples merely to fill space. A long weak item must remain weak, and a short\par
strong item may remain strong through one decisive or consequential clue.\par
\medskip
Expression modes:\par
- concrete\_episode: one observable event, choice, action, or reaction;\par
- preference\_or\_opinion: a natural first-person preference, choice, or view;\par
- self\_reflection: a recurring personal pattern noticed by the author without\par
  naming a construct.\par
\medskip
Every item must provide clean, directionally interpretable evidence for the\par
requested direction of the target construct. Do not include behavior that clearly\par
supports the other direction, and do not make another psychological construct\par
equally salient.\par
\medskip
An opposing-direction item must contain affirmative evidence for an opposing\par
pattern implied by the supplied definition. It must not rely only on omitting\par
aligned-direction behavior. A weak item must still be interpretable; weak means\par
subtle, not irrelevant.\par
\medskip
Write every item independently. Vary situations, syntax, openings, activities,\par
and details. Do not create mirrored, counterfactual, or template-like pairs.\par
\end{promptbox}

\begin{promptbox}[title={User prompt}]
Target construct: \{construct\_name\}\par
\medskip
Authoritative aligned-direction definition:\par
\{definition\_en\}\par
\medskip
"Aligned" means the direction described by the definition, and "opposing" means\par
the direction opposed to that definition. Strength alone controls the amount of\par
diagnostic evidence. Both directions require affirmative evidence in their\par
requested direction; mere absence of evidence for the other direction is not\par
sufficient. If the construct names a symptom, risk, or socially undesirable\par
tendency, do not reverse the directions for desirability.\par
\medskip
Use each requested context only as a setting and each expression\_mode only as\par
the form of writing. Across the batch, vary which behavioral or attitudinal\par
implication of the definition supplies the evidence when the definition permits\par
that variation.\par
\medskip
Generate exactly one independent text for every slot below. Each final text\par
must be within its inclusive word\_range. Verify every item's word count before\par
returning, and preserve every slot\_id exactly.\par
\medskip
\{slots\_json\}\par
\end{promptbox}

\paragraph{Comparison-based scoring and selection.}
Following ValueFlow \citep{kim2026valueflow}, we derive continuous scores from pairwise comparisons of the generated texts. Each candidate is compared with 20 distinct, randomly sampled candidates for the same construct, appearing ten times in each presentation position. DeepSeek-V4-Flash judges which text places its author further toward the direction specified by the construct definition, using temperature zero and returning an A/B choice for each comparison. The resulting 500{,}000 judgments are fitted separately for each construct with a Bradley--Terry model, using a comparison temperature of 0.15, scores bounded within $[-1,1]$, and a mean-zero constraint. Candidate selection considers score uncertainty, agreement with the comparison judgments, stability when scores are refitted on disjoint subsets of comparisons, and consistency with the requested direction. We also remove irrelevant texts, explicit psychological labels and duplicate narratives. We then sample across score bins to cover the full range of construct levels while maintaining diversity in context, expression form and length, retaining 50, 100 or 200 narratives per construct (15{,}250 measurement-training records in total) with their fitted scores as supervision targets. The comparison system and user prompts are given below.

\begin{promptbox}[title={System prompt}]
You are a careful psychometric pairwise rater.\par
\medskip
For each comparison, decide which text places its author relatively further\par
toward the direction specified by the supplied definition.\par
\medskip
Neither side is designated as the positive or negative example. Do not assume\par
that either side must strongly express the construct. A text may provide\par
definition-aligned evidence, definition-opposing evidence, or no relevant\par
evidence.\par
\medskip
Base the comparison only on construct-relevant evidence expressed in the text.\par
Do not use writing quality, topic, general sentiment, moral approval, social\par
desirability, or stereotypes as substitutes for the construct. For constructs\par
that describe symptoms, risks, or undesirable tendencies, the\par
definition-aligned direction remains the defined tendency, not the more\par
desirable person.\par
\medskip
Treat a text with no construct-relevant evidence as neutral for this comparison. Therefore,\par
definition-aligned evidence ranks above neutral, and neutral ranks above\par
definition-opposing evidence. These directions do not indicate whether the\par
content is positive, negative, desirable, or undesirable. Use one ordered\par
continuum: stronger opposing evidence < weaker opposing evidence < neutral <\par
weaker aligned evidence < stronger aligned evidence. Always select the text\par
farther toward the definition-aligned end of this continuum. Thus, when both\par
texts contain opposing evidence, select the one with weaker opposing evidence;\par
when both contain aligned evidence, select the one with stronger aligned\par
evidence.\par
\medskip
You must choose A or B. Do not return ties, confidence ratings, explanations,\par
or any text outside the required JSON object. Evaluate every comparison\par
independently.\par
\medskip
Return exactly:\par
\{"comparisons":[\{"pair\_id":"...","winner":"A"\}]\}\par
\medskip
Return one result for every pair\_id in the original order. Preserve every\par
pair\_id exactly and use only "A" or "B" as winner.\par
\end{promptbox}

\begin{promptbox}[title={User prompt}]
Target construct: \{construct\_name\}\par
\medskip
Authoritative aligned-direction definition:\par
\{definition\_en\}\par
\medskip
Comparisons:\par
\{comparisons\_json\}\par
\medskip
Return the required JSON object only.\par
\end{promptbox}

\subsection{Data synthesis for prediction training}\label{app:prediction-data-construction}

For prediction tasks derived from unlabelled personal writing, we use DeepSeek-V3.2 to extract discrete observations from individual texts. The extraction prompt asks for first-person statements about stable habits, interests, emotional tendencies, values or personality tendencies, each supported by an explicit quotation from the source text. We retain the extracted statements and pool them by author, together with available demographic descriptions. Approximately one quarter of the statements are randomly withheld, while the remaining records form the prediction input. Each withheld statement is then submitted separately to the same model to generate a four-option question and its reference answer. The prompt asks for either a fine-grained distinction between related behaviours or an abstract characterization of the underlying tendency, together with three plausible distractors. We parse the generated question, shuffle its options and update the reference answer accordingly. 
For prediction tasks, \chiron{} receives the remaining records and generated questions, with the withheld statements excluded from its input. The extraction and question-generation prompts are given below.

\begin{promptbox}[title={Observation extraction}]
You are a research assistant trained in computational social science.\par
\medskip
Your task is to extract ALL HIGH-CONFIDENCE personality-related features from a single blog post.\par
\medskip
---\par
\medskip
STRICT RULES:\par
\medskip
1. You must base your conclusions ONLY on explicit textual evidence.\par
2. You must NOT speculate or infer beyond what is clearly supported.\par
3. You may extract MULTIPLE features, but ONLY if each one is strongly supported.\par
4. If no feature is strongly supported, return an empty list.\par
5. Avoid clinical or medical diagnosis.\par
6. Be conservative: when in doubt, do NOT include the feature.\par
\medskip
---\par
\medskip
DEFINITION:\par
\medskip
A "feature" may include ONE of the following:\par
- A behavioral pattern (habit, routine)\par
- A stable interest\par
- A personality tendency (ONLY if clearly supported)\par
- A recurring emotional tendency\par
- A clear value or belief\par
\medskip
---\par
\medskip
OUTPUT FORMAT (STRICT):\par
\medskip
Return ONLY a JSON object:\par
\medskip
\{\par
  "features": [\par
    \{\par
      "feature": "A first-person declarative sentence (e.g., 'I tend to stay up late and struggle to wake up early.')",\par
      "category": "behavior | interest | personality | emotion | value",\par
      "evidence": ["short quote"],\par
      "confidence": "high"\par
    \}\par
  ]\par
\}\par
\medskip
OR:\par
\medskip
\{\par
  "features": []\par
\}\par
\medskip
---\par
\medskip
ADDITIONAL CONSTRAINTS:\par
\medskip
- Each feature MUST be written in FIRST PERSON ("I ...")\par
- Each MUST be a complete declarative sentence\par
- Each MUST reflect a stable tendency, not a one-time event\par
- Evidence must be a direct quote (max 20 words)\par
- ONLY include features with HIGH confidence\par
- Do NOT include weak or ambiguous traits\par
- Do NOT explain your reasoning\par
- Do NOT summarize the blog\par
\medskip
---\par
\medskip
DATA:\par
<BLOG\_HERE>\par
\end{promptbox}

\begin{promptbox}[title={Question generation}]
You are a data transformation assistant.\par
\medskip
Your task is to convert a first-person statement into a challenging multiple-choice question (4 options).\par
\medskip
Input:\par
A single first-person sentence describing a personal preference, habit, or trait.\par
\medskip
Output format (STRICT JSON, no extra text):\par
\{\par
  "question": "<a natural question in first-person perspective>",\par
  "option1": "<option>",\par
  "option2": "<option>",\par
  "option3": "<option>",\par
  "option4": "<option>",\par
  "groundtruth": <1, 2, 3, or 4>\par
\}\par
\medskip
Requirements:\par
\medskip
[Core Question Design]\par
1. The question must be written in first-person (use "I") and end with a question mark.\par
2. The question must NOT directly reuse key nouns or verbs from the input.\par
3. The question should target a higher-level preference, tendency, or behavioral pattern - NOT a direct restatement of the activity.\par
4. The question should require at least one step of inference (i.e., not directly answerable by keyword matching).\par
\medskip
[Two Allowed Difficulty Styles - choose ONE per example]\par
5. Use ONE of the following strategies to increase difficulty:\par
\medskip
   (A) Fine-grained distinction:\par
   - All four options belong to a similar broad domain but differ in subtle, meaningful ways.\par
   - Example dimensions:\par
     * types of activities (e.g., competitive sport vs leisure sport vs strategic game vs artistic hobby)\par
     * levels of accessibility (mass vs elite vs niche)\par
     * physical vs mental vs mixed engagement\par
   - Avoid overly similar options (e.g., basketball vs football).\par
   - Ensure distinctions are meaningful and inferable.\par
\medskip
   (B) Abstract categorization:\par
   - Map the input to a higher-level personality trait, cognitive style, or behavioral tendency.\par
   - Example dimensions:\par
     * structured vs spontaneous vs exploratory vs habitual\par
     * social vs independent vs reflective vs expressive\par
     * analytical vs creative vs physical vs routine-oriented\par
   - Options should be conceptual and require interpretation of the input.\par
\medskip
[Option Quality]\par
6. Exactly FOUR options must be provided.\par
7. Only ONE option is correct.\par
8. All options must:\par
   - be concise (short phrases)\par
   - be at a similar level of abstraction\par
   - be semantically plausible (no obviously wrong answers)\par
9. Avoid negations (e.g., "not doing something").\par
10. Ensure the incorrect options are strong distractors:\par
    - They should be reasonable interpretations but ultimately inconsistent with the input.\par
\medskip
[Balance \& Structure]\par
11. Randomly assign the correct answer to option1-option4 (ensure distribution balance over multiple samples).\par
12. Maintain symmetry:\par
    - Similar length\par
    - Similar structure\par
    - Similar abstraction level\par
\medskip
[Reasoning Constraint]\par
13. Internally perform:\par
    - Step 1: Identify the core feature of the input (e.g., physical, intellectual, social, routine-based, exploratory, etc.)\par
    - Step 2: Map it to either a fine-grained category OR an abstract trait\par
    - Step 3: Generate 3 high-quality distractors from nearby but distinct categories\par
\medskip
14. The final question must be solvable with reasonable inference, but not trivial.\par
\medskip
Output must be valid JSON (no comments, no trailing commas, no extra text).\par
\medskip
---\par
Example 1 (Fine-grained):\par
Input: I like playing basketball.\par
\medskip
Output:\par
\{\par
  "question": "What kind of activity do I tend to prefer?",\par
  "option1": "casual leisure activities",\par
  "option2": "strategic board-based challenges",\par
  "option3": "physically competitive team sports",\par
  "option4": "solo creative expression",\par
  "groundtruth": 3\par
\}\par
\medskip
---\par
Example 2 (Abstract):\par
Input: I enjoy reading books.\par
\medskip
Output:\par
\{\par
  "question": "What kind of personal tendency best describes me?",\par
  "option1": "seeking intellectual engagement",\par
  "option2": "pursuing high-energy stimulation",\par
  "option3": "favoring spontaneous exploration",\par
  "option4": "preferring social interaction",\par
  "groundtruth": 1\par
\}\par
\medskip
---\par
Input:\par
<input\_text>\par
\end{promptbox}

\subsection{Population synthesis}\label{app:synthesis-fit}

We synthesize respondent-level populations for the 63 training studies in \citet{cui2025largescale}, for which the human findings are available as aggregate statistics without the underlying individual responses. Archived LLM respondents from \citet{cui2025largescale} provide an initial pool of complete answer patterns. We reconstruct populations from this pool in three coarse-to-fine steps:

(1) \emph{Within-support reweighting:} We optimize candidate sampling weights via gradient-based optimization against the published human statistics and convert them into integer respondent counts while preserving each condition's original sample size. This step matches the target statistics as closely as possible without altering any individual item responses.

(2) \emph{Minimum-edit score projection:} Mode collapse in the source LLM pool often leaves tail or extreme responses unrepresented, preventing reweighting alone from reaching the reported human variance. We therefore use mixed-integer programming to determine how many respondents receive each attainable scale score to match the condition mean and standard deviation. For each respondent, we then project their answers onto the closest valid item-response vector (via exact subset-sum dynamic programming) that achieves the assigned score with the fewest edits to the source answers.

(3) \emph{Joint cross-condition refinement:} Finally, local search refines the population directly under the original study's statistical analysis procedure, evaluating paired experimental conditions jointly so that between-condition treatment effects ($d$) and significance outcomes are accurately recovered while preferring solutions with fewer source edits.

For a unified description of this procedure, we summarize the normalized fitting discrepancy and item-edit fraction as
\[
\begin{aligned}
E_{\mathrm{human}}(\mathcal{Y})
&=\frac{1}{K}\sum_{k=1}^{K}\frac{|\hat t_k(\mathcal{Y})-t_k|}{s_k},\\
D_{\mathrm{source}}(\mathcal{Y},Q)
&=\frac{1}{M}\sum_{p,i}\mathbf{1}[\tilde y_{pi}\neq q_{pi}].
\end{aligned}
\]
Here, $t_k$ and $\hat t_k$ are the published and reconstructed statistics, $s_k$ normalizes their scales, and $M$ counts the individual answers. The answers $\tilde y_{pi}$ and $q_{pi}$ belong to the reconstructed respondent and its paired source respondent selected by reweighting, respectively. Each changed item contributes equally, regardless of edit magnitude. Reweighting alone incurs no edit cost, even when it changes the response distribution. In the constrained formulation in \cref{eq:population-synthesis}, $\varepsilon$ specifies the fitting accuracy within which the item-edit fraction is minimized. The finalized respondent-level answers provide training targets for their corresponding experimental scenarios. \Cref{tab:synthesis-fit} summarizes agreement between the synthesized training populations and published human statistics. The 15 held-out evaluation studies are excluded from population construction.

\begin{table}[ht]
\centering
\caption{Agreement between synthesized training populations and human statistics.}
\label{tab:synthesis-fit}
\begin{tabular}{@{}lrr@{}}
\toprule
Metric & Eligible comparisons & Value \\
\midrule
$d$ MAE & 148 & 0.0120 \\
$d$ Pearson & 148 & 0.9990 \\
Direction accuracy & 218 & 0.9771 \\
Significance F1 & 319 & 0.9804 \\
Distribution NMAE & 175 & 0.0102 \\
\bottomrule
\end{tabular}
\end{table}

\subsection{Training mixture and optimization}\label{app:optimization}

The counts in \cref{tab:training-data-detailed} describe unique source records before assembling the joint training mixture. Expanding and repeating records by objective yields 220{,}096 prediction entries, 106{,}146 measurement entries (each measurement record included twice), and 115{,}955 population entries (each synthesized respondent included five times). We train on this mixture for two epochs using the optimization settings in \cref{tab:training-settings}.

\begin{table}[htbp]\centering
\caption{Optimization settings for joint training.}\label{tab:training-settings}
\begin{tabular}{@{}ll@{}}\toprule
Setting & Value \\\midrule
Adaptation & LoRA (applied to linear layers), rank 32, scaling 64 \\
LoRA dropout & 0 \\
Epochs & 2 \\
Optimizer & AdamW \\
Learning rate & $10^{-4}$ \\
Schedule & Cosine decay; 3\% warmup \\
Weight decay & 0 \\
Precision & BF16 \\
Maximum context length & 8{,}192 tokens \\
Hardware & Four NVIDIA A100 GPUs \\
Effective batch size & 12 records \\
Measurement loss weight & $\lambda_m=8$ \\
Population loss weight & $\lambda_p=1$ \\\bottomrule
\end{tabular}\end{table}

Construct directions $v_c$ in the measurement head are randomly initialized, with scales $a_c$ initialized to one and offsets $b_c$ initialized to zero.

\FloatBarrier

\section{Evaluation details and additional results}\label{app:evaluation}

\subsection{Psychological measurement}\label{app:measurement-details}

\subsubsection{Measurement evaluation protocol}\label{app:eval-measurement}

Across all measurement evaluations, \chiron{} and the four baseline models receive the same personal records. \chiron{} produces construct scores directly through its measurement head, whereas the baselines are prompted with the personal records together with the target construct's name and definition:

\begin{promptbox}[breakable=false,title={System prompt}]
You are a psychometric measurement model. Estimate one latent psychological construct from person-level records.\par
\medskip
Evidence rules:\par
1. Use only substantive evidence about the person inside <person\_records>...\allowbreak</person\_records>.\par
2. Do not use metadata, formatting, record order, or any provisional estimate outside that block as evidence.\par
3. Treat exact duplicate records as one observation; repetition does not create independent evidence.\par
4. Do not infer missing facts or rely on demographic stereotypes.\par
5. Records may be noisy or contradictory. Integrate the available evidence conservatively.
\end{promptbox}

\begin{promptbox}[breakable=false,title={User prompt}]
\{record\_context\}\par
<person\_records>\par
\{records\}\par
</person\_records>\par
\medskip
Target construct: \{construct\}\par
Definition: \{definition\}.\par
\medskip
\{task\_specific\_instructions\}
\end{promptbox}

Responses are parsed according to the requested output format and matched to their corresponding evaluation units.

\subsubsection{Convergent validity}\label{app:measurement-convergent}

In the five-dataset evaluation of convergent validity (\cref{sec:exp-humangold}), reference self-report scores are linearly mapped to $[-1,1]$ according to their source scales. For bounded ratings and questionnaire totals, we apply $2(x-l)/(u-l)-1$, using ranges of 1--9 for COVID-19 RW3D emotion ratings, 1--7 for Buechel empathic concern, 0--27 for PHQ-9 depression totals (PsihoRo and Student-SADD), and 0--21 for GAD-7 anxiety totals (PsihoRo). PANDORA personality percentiles $p\in[0,100]$ are mapped as $p/50-1$.

The five datasets yield 11 dataset--construct combinations covering nine constructs. For GPT-6 Sol, 127 requests that were refused after retries are excluded when computing its correlation and MAE. \Cref{tab:measurement-detailed-common} reports the dataset- and construct-level Pearson correlations and MAEs.

\begin{table}[H]
  \centering
  \caption{Convergent validity results (Pearson $r$ and MAE) against self-report assessments across datasets and constructs.}
  \label{tab:measurement-detailed-common}
  \resizebox{\linewidth}{!}{%
  \begin{tabular}{@{}llrrrrrrrrrr@{}}
  \toprule
  Dataset & Construct & \multicolumn{2}{c}{Chiron} & \multicolumn{2}{c}{GPT-6 Sol} & \multicolumn{2}{c}{Claude Opus 5.5} & \multicolumn{2}{c}{DeepSeek-V4.1-Flash} & \multicolumn{2}{c}{Qwen3.6-35B-A3B} \\
  \cmidrule(lr){3-12}
   & & $r\uparrow$ & MAE$\downarrow$ & $r\uparrow$ & MAE$\downarrow$ & $r\uparrow$ & MAE$\downarrow$ & $r\uparrow$ & MAE$\downarrow$ & $r\uparrow$ & MAE$\downarrow$ \\
  \midrule
  covid19\_real\_world\_worry\_rw3d & Anger & 0.4557 & 0.4094 & 0.3991 & 0.4994 & 0.3870 & 0.5444 & 0.3648 & 0.5036 & 0.3458 & 0.5720 \\
  covid19\_real\_world\_worry\_rw3d & anxiety & 0.6386 & 0.3243 & 0.6367 & 0.3270 & 0.5979 & 0.3384 & 0.6181 & 0.3318 & 0.5454 & 0.3801 \\
  empathic\_reactions\_buechel2018 & Empathic Concern & 0.4728 & 0.4296 & 0.3321 & 0.5916 & 0.3943 & 0.5878 & 0.3924 & 0.5534 & 0.3627 & 0.5987 \\
  pandora & agreeableness & 0.2507 & 0.5346 & 0.2951 & 0.5644 & 0.2898 & 0.5461 & 0.2982 & 0.5428 & 0.2559 & 0.5428 \\
  pandora & conscientiousness & 0.2662 & 0.5256 & 0.1924 & 0.6020 & 0.1271 & 0.5811 & 0.1484 & 0.5677 & 0.1332 & 0.5852 \\
  pandora & extraversion & 0.3395 & 0.4833 & 0.3142 & 0.5986 & 0.3249 & 0.5637 & 0.2996 & 0.5383 & 0.2269 & 0.5496 \\
  pandora & neuroticism & 0.2998 & 0.5466 & 0.2539 & 0.5454 & 0.2994 & 0.5461 & 0.2321 & 0.5486 & 0.2386 & 0.5561 \\
  pandora & openness & 0.2681 & 0.4537 & 0.2288 & 0.4425 & 0.1645 & 0.4551 & 0.1561 & 0.4547 & 0.1799 & 0.4730 \\
  psihoro\_romanian\_phq9\_gad7 & anxiety & 0.3443 & 0.3498 & 0.3694 & 0.7204 & 0.2962 & 0.7943 & 0.3633 & 0.7388 & 0.3720 & 0.7791 \\
  psihoro\_romanian\_phq9\_gad7 & depression & 0.3577 & 0.2639 & 0.4401 & 0.5332 & 0.3051 & 0.6563 & 0.2535 & 0.6582 & 0.3427 & 0.6394 \\
  student\_sadd\_phq9 & depression & 0.2958 & 0.3785 & 0.2545 & 0.3934 & 0.3037 & 0.3915 & 0.1794 & 0.4572 & 0.2894 & 0.4064 \\
  \midrule
  Mean across datasets & five datasets & 0.3903 & 0.3981 & 0.3532 & 0.5151 & 0.3464 & 0.5369 & 0.3197 & 0.5315 & 0.3324 & 0.5464 \\
  \bottomrule
  \end{tabular}%
  }
\end{table}

\subsubsection{Nomological validity}\label{app:eval-kg}

\paragraph{Construct relationships from the literature.}\label{app:kg-construction}

We construct the literature reference benchmark from a large corpus of published psychometric research articles. Specially tailored LLM agents first parse these articles to identify psychological construct entities and their reported empirical relationships, followed by manual review to verify the extracted records. From this collection, we select entries that report Pearson correlation coefficients and map their construct entities onto \chiron{}'s 109-construct inventory. Construct alignment relies on scale names and common aliases, consulting construct definitions, item wording, and surrounding article context whenever the correspondence is ambiguous. Across the full inventory, this mapping yields 346 construct pairs spanning 92 constructs. For each pair, we aggregate multiple literature findings by taking the median reported correlation coefficient, and retain pairs supported by at least two empirical entries. This filtering produces the final benchmark of 150 reference relationships across 60 constructs, which is evaluated across all three datasets.

\paragraph{Evaluation procedure and results.}

We evaluate these 150 relationships on three datasets with distinct record formats: PAN15 \citep{rangel2015pan15} (577 author-profiling participants), Essays \citep{pennebaker1999linguistic} (2{,}467 personal essay authors), and ESS09 \citep{ess2018round9} (4{,}921 survey respondents). Within each dataset, \chiron{} infers all 109 construct scores for each individual, and we compute the Pearson correlation across individuals for each of the 150 construct pairs. We then compare this 150-element correlation vector against the literature reference vector using Pearson correlation, sign agreement, and MAE (\cref{tab:kg-alignment}).

\begin{table}[htbp]\centering
\caption{Construct-relationship alignment with literature-derived correlations across 150 construct pairs (60 constructs) in each dataset. Pearson $r$, sign agreement, and MAE compare the vector of correlations among \chiron{}'s inferred construct scores with the literature median correlations.}
\label{tab:kg-alignment}
\begin{tabular}{@{}lrrrr@{}}
\toprule
Dataset & People & Relation $r$ & Sign agreement & MAE \\
\midrule
PAN15 & 577 & 0.8160 & 92.00\% & 0.3004 \\
Essays & 2,467 & 0.8060 & 91.33\% & 0.3027 \\
ESS09 & 4,921 & 0.8191 & 93.33\% & 0.3538 \\
\bottomrule\end{tabular}
\end{table}
\FloatBarrier

\subsubsection{Measurement bias}\label{app:measurement-bias}

\paragraph{Score dispersion.}
To test whether models compress inferred scores into a narrower range than human self-reports, we compare the dispersion of model predictions against self-reported Big Five personality percentiles on PANDORA \citep{gjurkovic2021pandora}. For each trait, we restrict the analysis to targets with valid scores from all five models, yielding 677 targets for agreeableness, 677 for conscientiousness, 676 for extraversion, 678 for neuroticism, and 677 for openness. Self-report percentiles are linearly mapped to $[-1,1]$ as described in \cref{app:measurement-convergent}, whereas model scores are evaluated directly without fitting, rescaling, or clamping (retaining one out-of-range \chiron{} score on neuroticism). For each model $m$ and trait $t$, we measure relative dispersion via the standard deviation ratio $R_{m,t}=\operatorname{SD}(\hat{y}_{m,t})/\operatorname{SD}(y_t)$, where $\hat{y}_{m,t}$ and $y_t$ denote the model scores and self-reports across the matched targets, and average $R_{m,t}$ with equal weight across the five traits. A ratio of $1$ indicates matching dispersion, values below $1$ indicate score compression, and values above $1$ indicate score expansion. Because $R_{m,t}$ compares marginal spread across individuals, it is invariant to mean shifts and distinct from individual-level accuracy. Trait-level descriptive statistics and SD ratios are reported in \cref{tab:pandora-score-distributions}.

\begin{longtable}{@{}llrrrrr@{}}
\caption{Score distributions and dispersion ratios across Big Five traits on PANDORA.}\label{tab:pandora-score-distributions}\\
\toprule Trait & Score source & $n$ & Mean & SD & SD ratio & MAE \\ \midrule
\endfirsthead
\toprule Trait & Score source & $n$ & Mean & SD & SD ratio & MAE \\ \midrule
\endhead
Agreeableness & Self-report & 677 & -0.159 & 0.618 & 1.000 & -- \\
 & Chiron & 677 & -0.076 & 0.376 & 0.608 & 0.531 \\
 & GPT-6 Sol & 677 & 0.117 & 0.343 & 0.554 & 0.564 \\
 & Claude Opus 5.5 & 677 & 0.055 & 0.327 & 0.529 & 0.543 \\
 & DeepSeek-V4.1-Flash & 677 & 0.052 & 0.312 & 0.504 & 0.540 \\
 & Qwen3.6-35B-A3B & 677 & -0.029 & 0.406 & 0.657 & 0.538 \\
\midrule
Conscientiousness & Self-report & 677 & -0.216 & 0.606 & 1.000 & -- \\
 & Chiron & 677 & -0.071 & 0.350 & 0.577 & 0.525 \\
 & GPT-6 Sol & 677 & 0.134 & 0.188 & 0.310 & 0.602 \\
 & Claude Opus 5.5 & 677 & 0.056 & 0.210 & 0.346 & 0.584 \\
 & DeepSeek-V4.1-Flash & 677 & 0.017 & 0.193 & 0.318 & 0.571 \\
 & Qwen3.6-35B-A3B & 677 & 0.057 & 0.257 & 0.423 & 0.590 \\
\midrule
Extraversion & Self-report & 676 & -0.272 & 0.612 & 1.000 & -- \\
 & Chiron & 676 & -0.248 & 0.350 & 0.573 & 0.487 \\
 & GPT-6 Sol & 676 & 0.097 & 0.247 & 0.404 & 0.599 \\
 & Claude Opus 5.5 & 676 & 0.026 & 0.267 & 0.437 & 0.567 \\
 & DeepSeek-V4.1-Flash & 676 & -0.073 & 0.220 & 0.360 & 0.541 \\
 & Qwen3.6-35B-A3B & 676 & -0.096 & 0.295 & 0.482 & 0.553 \\
\midrule
Neuroticism & Self-report & 678 & -0.002 & 0.645 & 1.000 & -- \\
 & Chiron & 678 & -0.099 & 0.367 & 0.569 & 0.547 \\
 & GPT-6 Sol & 678 & 0.098 & 0.251 & 0.388 & 0.545 \\
 & Claude Opus 5.5 & 678 & 0.163 & 0.210 & 0.325 & 0.546 \\
 & DeepSeek-V4.1-Flash & 678 & 0.047 & 0.205 & 0.318 & 0.549 \\
 & Qwen3.6-35B-A3B & 678 & 0.156 & 0.273 & 0.423 & 0.555 \\
\midrule
Openness & Self-report & 677 & 0.272 & 0.552 & 1.000 & -- \\
 & Chiron & 677 & 0.254 & 0.297 & 0.538 & 0.453 \\
 & GPT-6 Sol & 677 & 0.420 & 0.224 & 0.405 & 0.443 \\
 & Claude Opus 5.5 & 677 & 0.327 & 0.225 & 0.408 & 0.455 \\
 & DeepSeek-V4.1-Flash & 677 & 0.354 & 0.229 & 0.416 & 0.454 \\
 & Qwen3.6-35B-A3B & 677 & 0.336 & 0.346 & 0.627 & 0.473 \\
\bottomrule
\end{longtable}

\paragraph{Irrelevant context.}
To test whether construct scores remain invariant to construct-irrelevant context, we construct six-statement inputs that combine three target items measuring a given construct with three filler items drawn from other constructs. The filler items are assigned midpoint responses on their native scales (corresponding to a normalized score of zero), and all item--response pairs are rendered as natural-language statements using fixed templates (for example, ``It is neither accurate nor inaccurate to say that I keep my cool''). We then replace the three fillers with a different set of midpoint filler statements while keeping the target statements and their positions unchanged. Across 543 evaluation units spanning the 109 constructs (up to five units per construct), we score the original and modified inputs independently, compute the mean absolute score change within each construct, and average across constructs.

\paragraph{Evidence order.}
Using the same six-statement input structure (three target items and three midpoint fillers), we test whether scores depend on record ordering under two conditions: one with all three target responses indicating a low construct level, and another with all three indicating a high level, both accompanied by the same fillers. For each of the 540 units, we reverse the order of the six statements in both the low- and high-evidence conditions, calculate the absolute score change caused by reversal in each condition, and take the mean of the two changes. As above, results are summarized within constructs and then averaged across constructs.

Results for irrelevant context and evidence order are reported in \cref{tab:measurement-bias}.

\begin{table}[tb]\centering
\caption{Measurement-bias diagnostics across models. Lower values indicate smaller score changes.}\label{tab:measurement-bias}
\resizebox{\linewidth}{!}{%
\begin{tabular}{@{}lrrrrrr@{}}
\toprule
Diagnostic & $n$ & Chiron & GPT-6 Sol & Claude Opus 5.5 & DeepSeek-V4.1-Flash & Qwen3.6-35B-A3B \\
\midrule
Irrelevant context & 543 & 0.0238 & 0.0433 & 0.0560 & 0.0730 & 0.0850 \\
Evidence order & 540 & 0.0270 & 0.0433 & 0.0655 & 0.0684 & 0.1139 \\
\bottomrule\end{tabular}%
}
\end{table}
\FloatBarrier

\subsubsection{Experimental manipulation}\label{app:measurement-interventions}

All three intervention settings follow the same experimental manipulation logic: we hold a background context fixed and systematically manipulate construct-relevant evidence by replacing low-level evidence for a target construct with high-level evidence, testing whether the inferred construct score increases accordingly.

\paragraph{Graded interventions on questionnaires and narratives.}
For each of the 109 constructs, we construct four-level input series (six statements per input) by pairing three target statements with three fixed neutral background statements. Starting from Level~1 (all three target statements at the low level), we successively replace one, two and all three low-level target statements with their high-level counterparts (Levels~2--4), keeping the background statements and positions unchanged:
\begin{itemize}[leftmargin=1.5em,itemsep=2pt,topsep=2pt]
\item \emph{Questionnaire evidence} ($n=541$ series; up to five per construct): each target statement is a naturalized questionnaire item whose response is flipped from its low to its high endpoint (for example, changing ``I know how to captivate people'' from ``Very inaccurate'' to ``Very accurate''), producing four levels with mean target scores of $-1$, $-1/3$, $1/3$ and $1$. The three background statements are midpoint filler items from other constructs (\cref{app:measurement-bias}).
\item \emph{Narrative evidence} ($n=325$ series; up to three per construct): each low--high pair consists of two held-out synthetic narratives (\cref{sec:method-measurement}) with negative and positive Bradley--Terry scores, matched in scenario, writing style, length and absolute intensity (for example, preferring a quiet evening alone versus seeking out a friend's company for extraversion). The three background statements describe routine daily activities and are shared across constructs.
\end{itemize}

\paragraph{Interventions on real personal records.}
To test whether a single piece of contrasting evidence shifts scores in the expected direction against a realistic personal history, we sample 200 personal-record backgrounds each from PANDORA, DASS and IPIP-FFM ($n=592$ evaluated units). Each background is assigned a target construct and one held-out low--high narrative pair from above. We compare two versions of the input—one with the low-level narrative inserted and one with the high-level narrative inserted at the exact same position (placed before the original records in half of the units and after them in the other half).

\paragraph{Evaluation metrics.}
Every input version is scored independently. For the four-level questionnaire and narrative series, \emph{stepwise monotonicity} is the proportion of the three adjacent steps ($\text{Level }1\to2\to3\to4$) in which the target score strictly increases, and \emph{direction accuracy} is the proportion of series in which Level~4 scores higher than Level~1. For real personal records, \emph{direction accuracy} is the proportion of units in which the input with the high-level narrative scores higher than the input with the low-level narrative. All metrics are computed within each construct and then averaged across constructs (\cref{tab:controlled-evidence}).

\begin{table}[H]\centering
\caption{Stepwise monotonicity and direction accuracy under controlled evidence interventions across models.}\label{tab:controlled-evidence}
\resizebox{\linewidth}{!}{%
\begin{tabular}{@{}llrrrrrr@{}}
\toprule
Metric & Evidence type & $n$ & Chiron & GPT-6 Sol & Claude Opus 5.5 & DeepSeek-V4.1-Flash & Qwen3.6-35B-A3B \\
\midrule
Stepwise monotonicity & Questionnaire evidence & 541 & 99.45\% & 93.97\% & 90.46\% & 84.50\% & 67.24\% \\
 & Narrative evidence & 325 & 94.70\% & 94.34\% & 87.26\% & 88.02\% & 67.99\% \\
\midrule
Direction accuracy & Questionnaire evidence & 541 & 100.00\% & 99.27\% & 98.35\% & 97.06\% & 96.51\% \\
 & Narrative evidence & 325 & 100.00\% & 99.08\% & 99.39\% & 99.69\% & 98.47\% \\
 & Real personal records & 592 & 93.82\% & 96.36\% & 88.36\% & 92.49\% & 87.57\% \\
\bottomrule\end{tabular}%
}
\end{table}

\FloatBarrier

\subsection{Individual behavioural prediction}\label{app:prediction-details}

\subsubsection{Input formats and prompts}\label{app:prompt-system}\label{app:prompt-prediction}

For behavioural prediction, \chiron{} uses the following shared system prompt:

\begin{promptbox}[breakable=false,title={System prompt}]
You are a psychometric measurement and prediction model.\par
You will receive person-level records, such as statements, behaviors, events, or attributes. Use these records as evidence to answer the prediction questions about that person.\par
Output only the final answer(s).
\end{promptbox}

Each behavioural prediction example consists of a short description of the record context, the visible personal records enclosed in \texttt{<person\_records>}...\texttt{</person\_records>}, and one or more multiple-choice target questions:
\begin{promptbox}[breakable=false,title={User prompt}]
\{record\_context\}\par
<person\_records>\par
\{observations\}\par
</person\_records>\par
Questions:\par
\{questions\}
\end{promptbox}
The model reads the personal records and questions as a single input and outputs the predicted answer token(s) in order, without repeating the questions or generating explanations. During training, the cross-entropy loss is computed only on the reference answer tokens. The shortened fictional examples below illustrate the three types of prediction targets.

\paragraph{Questionnaire response.}
The target is one of the response options for a held-out questionnaire item, while the visible records contain other answers from the same person:
\begin{promptbox}[breakable=false]
<person\_records>\par
I prefer to spend free time with a small group of close friends.\par
I usually plan tomorrow's tasks before going to bed.\par
</person\_records>\par
Question 1: How often does the person plan activities in advance?\par
A. Never\quad B. Sometimes\quad C. Often\par
Target output: C
\end{promptbox}

\paragraph{Trait category.}
For writing datasets with external categorical annotations, the question asks for the annotated category directly:
\begin{promptbox}[breakable=false]
<person\_records>\par
I keep a detailed schedule and check each task before moving on.\par
When plans change, I reorganize the list and continue working.\par
</person\_records>\par
Question 1: Select the conscientiousness category.\par
Labels: low\quad medium\quad high\par
Target output: high
\end{promptbox}

\paragraph{Classification or masked observation.}
For classification tasks or masked observation prediction derived from personal writing, the withheld observation is presented as a multiple-choice question:
\begin{promptbox}[breakable=false]
<person\_records>\par
I usually turn down guided tours and figure out routes on my own.\par
At work, I would rather test a new approach than follow a fixed template.\par
</person\_records>\par
Question 1: What tends to guide the person's choices when taking on a new project?\par
A. Independent exploration\quad B. External approval\quad C. Routine stability\quad D. Material reward\par
Target output: A
\end{promptbox}

When an input contains multiple questions, the model outputs one numbered answer per line in the order of the questions, allowing a single forward pass to combine heterogeneous question formats:
\begin{promptbox}[breakable=false]
Question 1: What tends to guide the person's choices when taking on a new project?\par
A. Independent exploration\quad B. External approval\quad C. Routine stability\quad D. Material reward\par
Question 2: Select the conscientiousness category.\par
Labels: low\quad medium\quad high\par
Question 3: How often does the person plan activities in advance?\par
A. Never\quad B. Sometimes\quad C. Often\par
Target output:\par
Q1: A\par
Q2: high\par
Q3: C
\end{promptbox}

\subsubsection{Evaluation protocol and detailed results}

In the 29-domain prediction evaluation (\cref{sec:exp-prediction}), \chiron{} and the four baselines (GPT-6 Sol, Claude Opus 5.5, DeepSeek-V4.1-Flash and Qwen3.6-35B-A3B) receive the same personal records and target questions. All models are evaluated at temperature zero with reasoning disabled. Domain-level prediction accuracies and standard errors across all 29 domains are reported in \cref{tab:prediction-detailed}.

\begin{table}[!p]
  \centering
  \caption{Domain-level behavioural prediction results across all 29 domains. Domain entries report mean prediction accuracy $\pm1$ SEM. For GPT-6 Sol, 60 refused requests after retries are excluded when computing its mean accuracy; malformed outputs (two Blog cases for Qwen3.6-35B-A3B) are scored as zero.}
  \label{tab:prediction-detailed}
  \resizebox{\linewidth}{!}{%
  \begin{tabular}{@{}lrrrrrr@{}}
  \toprule
  Domain & $n$ & Chiron & GPT-6 Sol & Claude Opus 5.5 & DeepSeek-V4.1-Flash & Qwen3.6-35B-A3B \\
  \midrule
  \multicolumn{7}{@{}l}{\textit{Held-out individuals: 25 domains represented in prediction training}} \\
  16PF & 200 & $\mathbf{0.4735}\pm 0.0139$ & $0.3226\pm 0.0120$ & $0.3070\pm 0.0126$ & $0.3099\pm 0.0128$ & $0.2952\pm 0.0120$ \\
  AS+SC+AD+DO & 50 & $\mathbf{0.5070}\pm 0.0281$ & $0.4317\pm 0.0252$ & $0.4429\pm 0.0267$ & $0.4203\pm 0.0282$ & $0.4168\pm 0.0258$ \\
  Big Five & 200 & $\mathbf{0.4711}\pm 0.0146$ & $0.4313\pm 0.0135$ & $0.4198\pm 0.0131$ & $0.4141\pm 0.0139$ & $0.4062\pm 0.0139$ \\
  Blog & 200 & $\mathbf{0.8866}\pm 0.0098$ & $0.6413\pm 0.0144$ & $0.6533\pm 0.0131$ & $0.6820\pm 0.0130$ & $0.6381\pm 0.0134$ \\
  CFCS & 200 & $\mathbf{0.4600}\pm 0.0195$ & $0.4338\pm 0.0197$ & $0.4462\pm 0.0195$ & $0.4338\pm 0.0192$ & $0.3850\pm 0.0199$ \\
  DASS & 200 & $\mathbf{0.5359}\pm 0.0160$ & $0.4690\pm 0.0163$ & $0.4688\pm 0.0159$ & $0.4679\pm 0.0156$ & $0.4526\pm 0.0163$ \\
  ECR & 200 & $\mathbf{0.5344}\pm 0.0157$ & $0.4884\pm 0.0147$ & $0.4763\pm 0.0144$ & $0.4583\pm 0.0142$ & $0.4143\pm 0.0154$ \\
  Essays Big Five & 123 & $\mathbf{0.6325}\pm 0.0224$ & $0.5902\pm 0.0208$ & $0.5772\pm 0.0192$ & $0.5496\pm 0.0207$ & $0.5577\pm 0.0198$ \\
  FTI & 200 & $\mathbf{0.5601}\pm 0.0147$ & $0.5214\pm 0.0136$ & $0.4969\pm 0.0134$ & $0.5146\pm 0.0140$ & $0.3884\pm 0.0145$ \\
  GRIT tweets & 21 & $0.2875\pm 0.0329$ & $\mathbf{0.2977}\pm 0.0531$ & $0.2898\pm 0.0437$ & $0.2807\pm 0.0513$ & $0.2205\pm 0.0398$ \\
  HBDS & 200 & $\mathbf{0.3912}\pm 0.0165$ & $0.2935\pm 0.0145$ & $0.2885\pm 0.0147$ & $0.2742\pm 0.0155$ & $0.3088\pm 0.0159$ \\
  HSNS+DD & 200 & $\mathbf{0.5054}\pm 0.0173$ & $0.3816\pm 0.0147$ & $0.3938\pm 0.0161$ & $0.3830\pm 0.0156$ & $0.3818\pm 0.0171$ \\
  HSQ & 54 & $\mathbf{0.2850}\pm 0.0231$ & $0.2306\pm 0.0247$ & $0.1713\pm 0.0231$ & $0.2260\pm 0.0196$ & $0.1581\pm 0.0232$ \\
  IPIP-FFM & 200 & $\mathbf{0.4945}\pm 0.0154$ & $0.4339\pm 0.0144$ & $0.4067\pm 0.0144$ & $0.4081\pm 0.0149$ & $0.4186\pm 0.0151$ \\
  NPAS & 200 & $\mathbf{0.4181}\pm 0.0153$ & $0.2672\pm 0.0117$ & $0.3040\pm 0.0126$ & $0.2555\pm 0.0124$ & $0.3012\pm 0.0139$ \\
  NPI & 200 & $\mathbf{0.7554}\pm 0.0130$ & $0.7455\pm 0.0135$ & $0.7467\pm 0.0138$ & $0.7184\pm 0.0147$ & $0.6560\pm 0.0167$ \\
  NR-6 & 111 & $\mathbf{0.4721}\pm 0.0209$ & $0.3604\pm 0.0198$ & $0.3387\pm 0.0212$ & $0.3081\pm 0.0196$ & $0.3586\pm 0.0225$ \\
  OSRI-44 & 200 & $\mathbf{0.4397}\pm 0.0157$ & $0.2949\pm 0.0135$ & $0.3124\pm 0.0135$ & $0.2130\pm 0.0125$ & $0.3118\pm 0.0137$ \\
  PANDORA & 71 & $\mathbf{0.3746}\pm 0.0254$ & $0.2479\pm 0.0258$ & $0.2141\pm 0.0264$ & $0.2366\pm 0.0267$ & $0.2507\pm 0.0233$ \\
  PWE & 67 & $\mathbf{0.4122}\pm 0.0238$ & $0.3872\pm 0.0211$ & $0.3712\pm 0.0192$ & $0.3267\pm 0.0232$ & $0.3265\pm 0.0221$ \\
  RSE & 200 & $\mathbf{0.6383}\pm 0.0216$ & $0.6250\pm 0.0203$ & $0.5583\pm 0.0213$ & $0.6300\pm 0.0206$ & $0.5233\pm 0.0227$ \\
  SD3 & 200 & $\mathbf{0.4590}\pm 0.0166$ & $0.3818\pm 0.0152$ & $0.4073\pm 0.0149$ & $0.3545\pm 0.0138$ & $0.3461\pm 0.0146$ \\
  Self-esteem & 162 & $\mathbf{0.8827}\pm 0.0254$ & $0.4841\pm 0.0447$ & $0.2778\pm 0.0353$ & $0.3704\pm 0.0381$ & $0.3889\pm 0.0384$ \\
  Twitter ST-MHD & 108 & $\mathbf{0.4630}\pm 0.0482$ & $0.1731\pm 0.0373$ & $0.1944\pm 0.0383$ & $0.2315\pm 0.0408$ & $0.2037\pm 0.0389$ \\
  WVS & 200 & $\mathbf{0.7126}\pm 0.0133$ & $0.6984\pm 0.0125$ & $0.6717\pm 0.0126$ & $0.6739\pm 0.0132$ & $0.6258\pm 0.0135$ \\
  \midrule
  \multicolumn{7}{@{}l}{\textit{Held-out domains: four domains absent from prediction training}} \\
  HEXACO & 200 & $\mathbf{0.3616}\pm 0.0139$ & $0.3249\pm 0.0125$ & $0.3192\pm 0.0128$ & $0.2758\pm 0.0123$ & $0.2830\pm 0.0126$ \\
  MACH-IV & 200 & $\mathbf{0.3946}\pm 0.0150$ & $0.3609\pm 0.0142$ & $0.3243\pm 0.0130$ & $0.2622\pm 0.0126$ & $0.3302\pm 0.0153$ \\
  RIASEC & 200 & $\mathbf{0.4644}\pm 0.0146$ & $0.4558\pm 0.0145$ & $0.4248\pm 0.0152$ & $0.3900\pm 0.0145$ & $0.4255\pm 0.0142$ \\
  SAPA-2022 & 200 & $\mathbf{0.3432}\pm 0.0130$ & $0.3384\pm 0.0144$ & $0.3397\pm 0.0131$ & $0.3182\pm 0.0147$ & $0.2937\pm 0.0131$ \\
  \midrule
  25-domain mean & 3,967 & \textbf{0.5221} & 0.4253 & 0.4094 & 0.4056 & 0.3894 \\
  Four-domain mean & 800 & \textbf{0.3910} & 0.3700 & 0.3520 & 0.3115 & 0.3331 \\
  All-domain mean & 4,767 & \textbf{0.5040} & 0.4177 & 0.4015 & 0.3927 & 0.3816 \\
  \bottomrule
  \end{tabular}%
  }
\end{table}

\subsubsection{Detailed experimental setup for linking construct measurement to prediction}\label{app:measurement-prediction-linkage}

We test whether \chiron{}'s construct measurements and predicted questionnaire responses change together when evidence about a hypothetical person is varied (\cref{fig:prediction-case-study}e). The experiment covers all 109 constructs.

For each construct, we create records containing three fixed statements about everyday background and three responses to questionnaire items. The item responses describe different levels of the construct. For example, agreement with ``I follow a schedule'' provides evidence of higher conscientiousness. We make four versions of each record: at L1, all three responses indicate low construct levels; one response is switched to its high-level counterpart at each step, until all three indicate high levels at L4. For an item describing disorganized habits, disagreement supplies the high-conscientiousness response. 
For each of the 109 constructs, we combine five sets of background notes with four levels of construct-relevant evidence, yielding 2{,}180 input records.

Three other items about the same construct are reserved as prediction questions. They never appear in that construct's input records and remain fixed across its backgrounds and evidence levels. The model predicts how the hypothetical person would answer each question, giving 6{,}540 item predictions. Of the 327 construct-specific target questions, 262 use existing questionnaire items, and the other 65 are written from construct definitions when existing items are not sufficient.

During evaluation, the model returns a construct score using only the personal records. Each prediction question is then presented separately. We score its possible \texttt{Option N} answers and normalize their probabilities over the listed options. The expected response is the probability-weighted mean of the option numbers. We linearly rescale this expectation to $[-1,1]$, reversing items for which larger responses indicate less of the construct. Higher prediction scores therefore always indicate more of the target construct. We average the three transformed expectations to obtain one prediction score for the record. A transition between adjacent evidence levels counts as jointly consistent only when both the measured construct score and the prediction score strictly increase. We average these results over the 15 transitions per construct (five backgrounds, three transitions each), then equally across constructs.

\subsection{Population simulation}\label{app:population-details}

\subsubsection{Evaluation protocol and prompts}\label{app:population-coverage}\label{app:eval-population}\label{app:prompt-population}

We evaluate population simulation on 15 held-out studies comprising 45 experimental conditions and 4{,}277 planned virtual respondents per model (\cref{tab:data-population}). For each condition, \chiron{} uses the shared system prompt in \cref{app:prompt-system} and receives the experimental scenario and study items formatted as follows:

\begin{promptbox}[breakable=false,title={User prompt}]
Below are records describing one participant in an experimental scenario.\par
<person\_records>\par
\{experimental scenario and condition text\}\par
</person\_records>\par
\medskip
Questions:\par
Q1. \{study-specific question content\}\par
Please predict this participant's response to the following item:\par
\{study-specific item content\}\par
Option 1: \{option text\}\par
Option 2: \{option text\}\par
...\par
\medskip
Q2. \{study-specific question content\}\par
Please predict this participant's response to the following item:\par
\{study-specific item content\}\par
Option 1: \{option text\}\par
Option 2: \{option text\}\par
...
\end{promptbox}

\begin{promptbox}[breakable=false,title={Assistant target format}]
Q1: Option \{option\_id\}\par
Q2: Option \{option\_id\}\par
...
\end{promptbox}

For each condition, we sample as many independent completions from \chiron{} as the human condition sample size using temperature one and top-$p$ one. Baseline models (GPT-6 Sol, Claude Opus 5.5, DeepSeek-V4.1-Flash and Qwen3.6-35B-A3B) follow the scenario-based replication protocol of \citet{cui2025largescale}.

\subsubsection{Detailed results}\label{app:population-study-results}
Following \citet{cui2025largescale}, we apply each study's original scoring procedure and summarize agreement between simulated and human outcomes across five metrics: Cohen's $d$ MAE and Pearson correlation (comparing human and simulated standardized effect sizes), effect direction accuracy (whether simulated and human effects share the same sign), statistical significance F1 (treating human $p<0.05$ significance decisions as reference labels), and response-distribution NMAE (measuring normalized errors in condition means and standard deviations).

Because not every study reports all statistics, the 15-study evaluation comprises 25 eligible effect-size pairs (from eight studies), 42 direction comparisons, 53 significance comparisons, and 34 distribution comparisons. Aggregate results across all five models are reported in \cref{tab:population}.

\begin{table}[tb]\centering
\caption{Aggregate experimental results of population simulation.}\label{tab:population}
\begin{tabular}{@{}lrrrrr@{}}\toprule
Model & $d$ MAE$\downarrow$ & $d$ Pearson$\uparrow$ & Direction$\uparrow$ & Sig. F1$\uparrow$ & Dist. NMAE$\downarrow$ \\\midrule
Chiron & 0.2239 & 0.8729 & 0.9048 & 0.6780 & 0.0755 \\
GPT-6 Sol & 3.8951 & 0.5397 & 0.7857 & 0.7750 & 0.1638 \\
Claude Opus 5.5 & 1.9565 & 0.7032 & 0.8333 & 0.7407 & 0.1402 \\
DeepSeek-V4.1-Flash & 1.2542 & 0.6787 & 0.8095 & 0.7436 & 0.1255 \\
Qwen3.6-35B-A3B & 0.7977 & 0.6730 & 0.8095 & 0.7848 & 0.1248 \\
\bottomrule\end{tabular}\end{table}

\FloatBarrier

\end{document}